\pdfoutput=1
\documentclass[11pt]{article}

\usepackage[]{acl}
\usepackage{times}
\usepackage{latexsym}

\usepackage[T1]{fontenc}
\usepackage[utf8]{inputenc}

\usepackage{microtype}

\usepackage[ruled]{algorithm2e}
\usepackage{booktabs}
\usepackage{times}
\usepackage{latexsym}
\usepackage{graphicx}
\usepackage{multirow}
\usepackage{multicol}
\usepackage{hyperref}
\usepackage{cite}
\usepackage{makecell}
\usepackage[flushleft]{threeparttable}
\usepackage{amsmath,amssymb,amsfonts}
\usepackage{scalerel,xparse}

\newcolumntype{H}{>{\setbox0=\hbox\bgroup}c<{\egroup}@{}}

\newcommand{\treelogo}{\raisebox{5pt}{\includegraphics[scale=0.050]{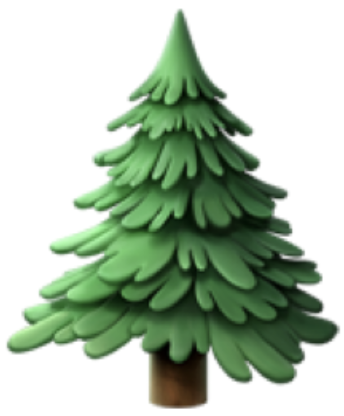}}}
\newcommand{\amazon}{\raisebox{3.5pt}{\includegraphics[scale=0.028]{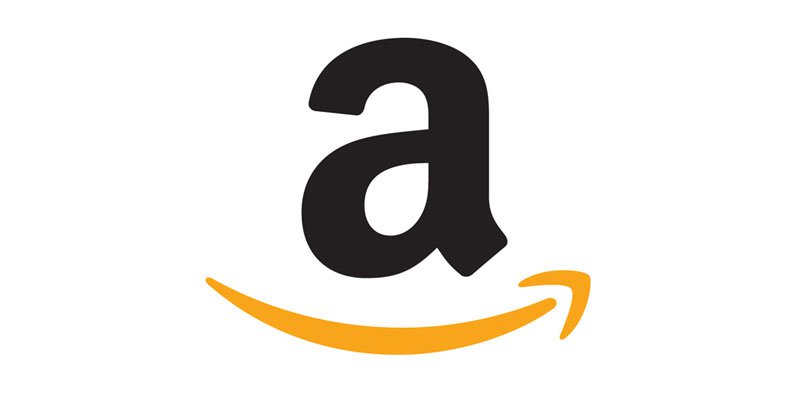}}}
\newcommand{\gtlogo}{\raisebox{3.4pt}{\includegraphics[scale=0.025]{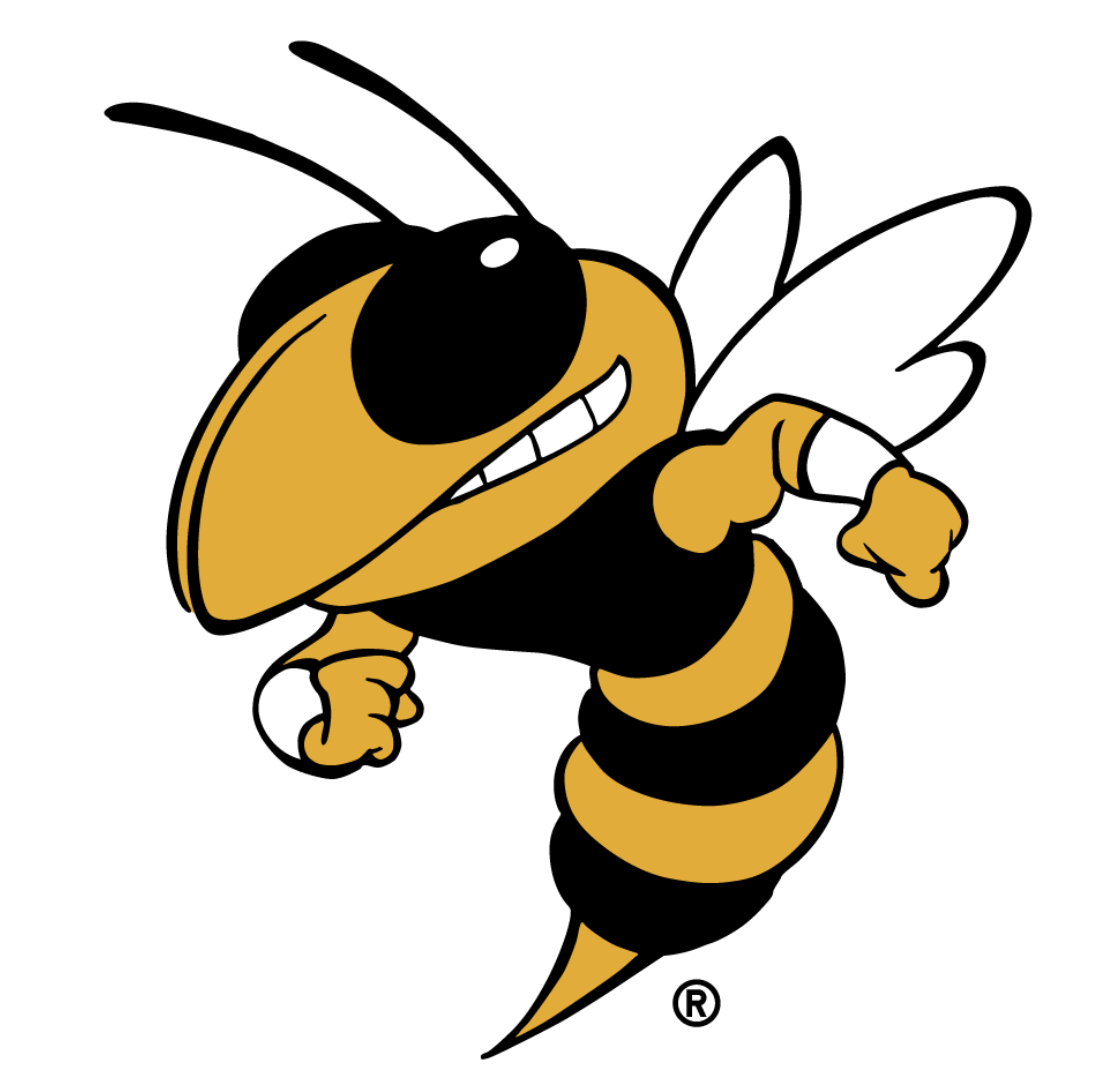}}}

\newcommand{\fire}{\raisebox{5pt}{\includegraphics[scale=0.15]{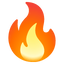}}}

\usepackage{pifont}

\usepackage{soul} 
\usepackage[skins]{tcolorbox} 
\newcommand{\drop}[1]{#1{\color[HTML]{CB4335}$^-$}}
\newcommand{\up}[1]{#1{\color[HTML]{2E86C1}$^+$}}

\newcommand{\gold}[1]{{\color[HTML]{d4af37}#1}}
\definecolor{valbest}{HTML}{d9ead3}
\newcommand{\valbest}[1]{\colorbox{valbest}{#1}}
\definecolor{valgood}{HTML}{d0e0e3}
\newcommand{\valgood}[1]{\colorbox{valgood}{#1}}
\definecolor{valbad}{HTML}{ead1dc}
\newcommand{\valbad}[1]{\colorbox{valbad}{#1}}
\usepackage{tipa}
\usepackage{dblfloatfix}
\usepackage{amsmath}
\usepackage{graphicx}
\usepackage{array,booktabs,ragged2e}
\usepackage{multirow}
\newcolumntype{R}[1]{>{\RaggedLeft\arraybackslash}p{#1}}
\newcolumntype{L}[1]{>{\RaggedRight\arraybackslash}p{#1}}
\newcolumntype{M}[1]{>{\centering\arraybackslash}m{#1}}
\usepackage{xcolor}
\usepackage{enumitem}
\usepackage{tipa}

\newcommand{\mval}[1]{Multi-VALUE#1}
\newcommand{\numFeatures}{189}
\newcommand{\numDialects}{50}

\newcommand{\averageDialectCoverage}{86.6\%}
\newcommand{\ewave}[1]{\href{https://ewave-atlas.org/parameters/#1}{#1}}
\newcommand{\hide}[1]{}

\newcommand\coauth{\fire}
\newcommand{\gt}{\gtlogo}
\newcommand{\amz}{\amazon}
\newcommand{\stanf}{\treelogo}
\newcommand\blfootnote[1]{%
  \begingroup
  \renewcommand\thefootnote{}\footnote{#1}%
  \addtocounter{footnote}{-1}%
  \endgroup
}

\newtcolorbox{marker}[2][]{enhanced,nobeforeafter,tcbox raise base,boxrule=0.4pt,top=0mm,bottom=0mm,
  right=0mm,left=4mm,arc=1pt,boxsep=2pt,before upper={\vphantom{dlg}},
  colframe=green!50!black,coltext=green!25!black,colback=green!10!white,
  overlay={\begin{tcbclipinterior}\fill[green!75!blue!50!white] (frame.south west)
    rectangle node[text=white,font=\sffamily\bfseries\small] {#2} ([xshift=4mm]frame.north west);\end{tcbclipinterior}},#1}

\newtcbox{\one}{enhanced,nobeforeafter,tcbox raise base,boxrule=0.4pt,top=0mm,bottom=0mm,
  right=0mm,left=4mm,arc=1pt,boxsep=2pt,before upper={\vphantom{dlg}},
  colframe=green!50!black,coltext=green!25!black,colback=green!10!white,
  overlay={\begin{tcbclipinterior}\fill[green!75!blue!50!white] (frame.south west)
    rectangle node[text=white,font=\sffamily\bfseries\small] {1} ([xshift=4mm]frame.north west);\end{tcbclipinterior}}}

\newtcbox{\two}{enhanced,nobeforeafter,tcbox raise base,boxrule=0.4pt,top=0mm,bottom=0mm,
  right=0mm,left=4mm,arc=1pt,boxsep=2pt,before upper={\vphantom{dlg}},
  colframe=green!50!black,coltext=green!25!black,colback=green!10!white,
  overlay={\begin{tcbclipinterior}\fill[green!75!blue!50!white] (frame.south west)
    rectangle node[text=white,font=\sffamily\bfseries\small] {2} ([xshift=4mm]frame.north west);\end{tcbclipinterior}}}

\newtcbox{\three}{enhanced,nobeforeafter,tcbox raise base,boxrule=0.4pt,top=0mm,bottom=0mm,
  right=0mm,left=4mm,arc=1pt,boxsep=2pt,before upper={\vphantom{dlg}},
  colframe=green!50!black,coltext=green!25!black,colback=green!10!white,
  overlay={\begin{tcbclipinterior}\fill[green!75!blue!50!white] (frame.south west)
    rectangle node[text=white,font=\sffamily\bfseries\small] {3} ([xshift=4mm]frame.north west);\end{tcbclipinterior}}}

\newtcbox{\four}{enhanced,nobeforeafter,tcbox raise base,boxrule=0.4pt,top=0mm,bottom=0mm,
  right=0mm,left=4mm,arc=1pt,boxsep=2pt,before upper={\vphantom{dlg}},
  colframe=green!50!black,coltext=green!25!black,colback=green!10!white,
  overlay={\begin{tcbclipinterior}\fill[green!75!blue!50!white] (frame.south west)
    rectangle node[text=white,font=\sffamily\bfseries\small] {4} ([xshift=4mm]frame.north west);\end{tcbclipinterior}}}

\newtcbox{\five}{enhanced,nobeforeafter,tcbox raise base,boxrule=0.4pt,top=0mm,bottom=0mm,
  right=0mm,left=4mm,arc=1pt,boxsep=2pt,before upper={\vphantom{dlg}},
  colframe=green!50!black,coltext=green!25!black,colback=green!10!white,
  overlay={\begin{tcbclipinterior}\fill[green!75!blue!50!white] (frame.south west)
    rectangle node[text=white,font=\sffamily\bfseries\small] {5} ([xshift=4mm]frame.north west);\end{tcbclipinterior}}}

\title{\mval{:} A Framework for Cross-Dialectal English NLP}

\author{Caleb Ziems \coauth \stanf \hspace{1.5em}
        William Held \coauth \gt \hspace{1.5em}
        Jingfeng Yang\amz\\
        \textbf{Jwala Dhamala}\amz \hspace{1.5em}
        \textbf{Rahul Gupta}\amz \hspace{1.5em}
        \textbf{Diyi Yang}\stanf \hspace{1.5em} \\
        \stanf Stanford University, \gt Georgia Institute of Technology, \amz Amazon\\
        \texttt{\{cziems, diyiy\}@stanford.edu}, \texttt{\{wheld3\}@gatech.edu},\\
        \texttt{\{jddhamal, yjfllpyym, gupra\}@amazon.com}
}

\begin{document}
    \maketitle

\begin{abstract}
Dialect differences caused by regional, social, and economic factors cause performance discrepancies for many groups of language technology users. Inclusive and equitable language technology must critically be dialect invariant, meaning that performance remains constant over dialectal shifts. Current systems often fall short of this ideal since they are designed and tested on a single dialect: Standard American English (SAE). We introduce a suite of resources for evaluating and achieving English dialect invariance. The resource is called \mval{}, a controllable rule-based translation system spanning \numDialects{} English dialects and \numFeatures{} unique linguistic features. \mval{} maps SAE to synthetic forms of each dialect. 
First, we use this system to stress tests question answering, machine translation, and semantic parsing. Stress tests reveal significant performance disparities for leading models on non-standard dialects. Second, we use this system as a data augmentation technique to improve the dialect robustness of existing systems. Finally, we partner with native speakers of Chicano and Indian English to release new gold-standard variants of the popular CoQA task. To execute the transformation code, run model checkpoints, and download both synthetic and gold-standard dialectal benchmark datasets, see \url{http://value-nlp.org/}.
\blfootnote{\coauth Equal contribution.}

\end{abstract}
  
\section{Introduction}
\label{sec:intro}

\begin{figure*}[t]
\centering
    \includegraphics[width=0.99\textwidth]{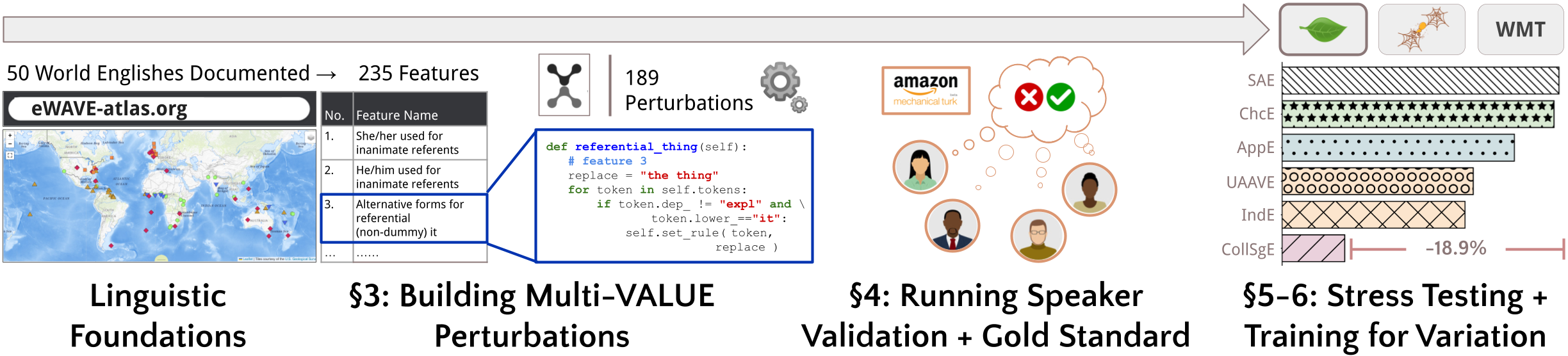}
    \caption{\textbf{The \mval{} pipeline} is grounded in a set of \numFeatures{} linguistic structures from the dialectology literature (\S\ref{sec:intro}). For each structure, we write a perturbation rule to inject it into text (\S\ref{sec:dialect_transformations}). By partnering with native speakers, we validate perturbations and build both synthetic and gold standard benchmarks (\S\ref{sec:validation_reliability}-\ref{sec:using_mval}). Finally, we use these resources in (1) stress testing supervised models to reveal dialect disparities and (2) fine-tuning these models on synthetic data to close the performance gap (\S\ref{sec:benchmarking_experiments}).}
    \label{fig:pipeline}
\end{figure*}

\begin{quote}
    {\small
    ``\textit{[Often, speakers] will not be hampered by the lack of language technology in their local language, but by the lack of support for their variety of the contact language.}'' \\\phantom{abc}--- \textbf{Steven Bird} (\citeyear{bird2022local})
    }
\end{quote}

Global contact languages like English will continue to have an outsized impact on commerce, economics, wellbeing, and equity worldwide. English, like any other language, is subject to variation across time~\citep{yang2000internal} and between speakers or speaker groups~\citep{eckert2017age,holmes2008handbook}. Rather than focusing on social status or political power~\citep{stewart1968sociolinguistic,chambers1998dialectology}, linguists define \textit{dialects} as descriptive sets of correlated \textit{features} common across a group of speakers~\citep{nerbonne2009data}. Current pre-training paradigms employ content filters that can exclude text in English dialects other than Standard American and British~\citep{gururangan2022whose}, which leads to performance gaps for other varieties. These discrepancies in Natural Language Processing (NLP) cause allocational harms for dialectal speakers in downstream applications~\citep{bender2021dangers}, making dialect robustness a critical need for fair and inclusive language technology. 

This disparity is clear in a growing body of empirical work on African American English \citep{ziems-etal-2022-value,halevy2021mitigating,blodgett2018twitter, jurgens-etal-2017-incorporating,kiritchenko-mohammad-2016-effect}. However, there does not yet exist a systematic exploration of robustness across multiple Englishes, nor of models' ability to transfer knowledge between varieties with similar features, as in multi-lingual NLP. We need new tools to benchmark and achieve dialect robustness.

We introduce \textbf{\mval{}}\footnote{Multi-VALUE is a \textbf{Multi}-dialectal \textbf{V}ern\textbf{A}cular \textbf{L}anguage \textbf{U}nderstanding \textbf{E}valuation framework (\url{value-nlp.org})} for English dialect robustness. Our feature-based approach leverages decades of field linguistics research to isolate grammatical constructions~\citep{demszky2021learning} that vary in \textit{regional} Englishes~\citep{labov1972language,eckert1989jocks,hovy2021importance}. \hide{across dialects from topical shifts found in data collected using non-linguistic criteria such as geolocation~\citep{blodgett2016demographic}.}
We focus on varieties that (1) are mutually intelligible with Standard American English (SAE); (2) share vocabulary with SAE; and (3) differ from SAE with respect to \textit{morphology} and \textit{syntax}. The third criterion defines the critical axis of variation. The first two criteria ensure that our definition of model robustness aligns with the human ability to understand other varieties. For example, creoles have their own unique vocabularies and are not easily understood by speakers of other Englishes \citep{contactlang1997creolesanddialects}; they are outside the scope of this study. 

First, we provide a controllable \textbf{(1)~rule-based translation system} for injecting up to \numFeatures{} features into SAE text. This will allow researchers and practitioners to build \textit{synthetic training data} plus on-demand \textit{dialect stress tests} for nearly any task. We stress test leading models for three challenging tasks and find statistically significant performance gaps. Second, we provide reliable \textbf{(2)~gold standard benchmarks} for the CoQA task in two widely-spoken varieties: Chicano and Indian English. We find that, by training models on synthetic data, we improve dialectal robustness. Third, we fine-tune and publish \textbf{(3)~dialect-robust models} on the HuggingFace Hub~\citep{wolf-etal-2020-transformers}, which can be used directly in downstream applications. Figure~\ref{fig:pipeline} demonstrates the full project pipeline.

We recognize five advantages in the \mval{} approach. Our system is
\begin{enumerate}[label=(\Alph*)]\setlength\itemsep{0em}
    \item \textbf{Interpretable:} supports systematic perturbation analyses
    \item \textbf{Flexible:} customized to align with new and evolving dialects by adjusting the \textit{density} of dialectal features, unlike fixed or static datasets.
    \item \textbf{Scalable:} allows users to mix and match tasks and dialects at scale without the need for costly human annotation.
    \item \textbf{Responsible:} vetted by native speakers to ensure gold standards and synthetic data are dependable for ongoing research.
    \item \textbf{Generalizable:} moves the field beyond single-dialect evaluation, which allows researchers to draw more transferrable findings about cross-dialectal NLP performance.
\end{enumerate}

\section{Related Work}

\paragraph{Dialect Disparity}
\label{subsec:lit_dialect_disparity}
is an issue of equity and fairness \citep{hovy2016social,gururangan2022whose,halevy2021mitigating,blodgett2017racial}. There is mounting evidence of dialect disparity in NLP. Hate speech classifiers have known biases against African American English \citep{davidson2019racial,mozafari2020hate,rios2020fuzze,sap2019risk,zhou2021challenges}. Text from regions with a predominantly Black population are more likely to be classified as hate speech \citep{mozafari2020hate,sap2019risk,davidson2019racial}. AAVE performance gaps have also been found across a wide range of core NLP tasks like NLI \citep{ziems-etal-2022-value}, dependency parsing and POS tagging \citep{blodgett2018twitter, jorgensen2015challenges}, plus downstream applications \hide{like influenza detection} \citep{lwowski2021risk}. Still, there does not \hide{yet} exist a systematic study on cross-dialectal model performance. We aim to fill this gap, expanding the VernAcular Language Understanding Evaluation (VALUE) framework of \citet{ziems-etal-2022-value}. Where VALUE established a uni-dialectal evaluation harness with 11 perturbation rules, Multi-VALUE now supports multi-dialectal evaluation with \numFeatures{} different perturbations across \numDialects{} English dialects. Our empirical study on dialect disparity is also more expansive than prior work as we consider three separate domains: QA, MT, and semantic parsing.

\paragraph{Multilingual NLP}
\label{related_work:multilingual}
studies how to learn common structures that transfer across languages. \hide{Since dialectal variation is systematic,} These strategies may also yield benefits in multi-dialectal settings. \hide{especially in cases of mutual intelligibility.}
\hide{Multilingual training and evaluation have led to large advances in many languages.} Massively multilingual models~\citep{mbert, conneau2020unsupervised, mbart, mt5-xue} exploit the commonalities between many languages at once, rather than merely achieving pairwise transfer~\citep{lin-etal-2019-choosing}. Additionally, benchmarking across multiple languages can reveal language discrepancies at the modeling level, even without language-specific feature engineering or training data~\citep{bender2011achieving, ravfogel-etal-2018-lstm, ahmad-etal-2019-difficulties, tsarfaty-etal-2020-spmrl}. \mval{} aims to bring these advantages\hide{identified from multilingual NLP} to the study of English dialects.

\section{\mval{} Perturbations}
\label{sec:dialect_transformations}

There is a clear need for dialect robustness (\S\ref{subsec:lit_dialect_disparity}). The challenge is that language is subject to \textit{variation} and \textit{change}. This means speakers can contextually modulate the density of features in their grammar, and over time, speakers adopt different features. Shifting language can quickly antiquate training and testing data, and updating such resources can be costly and time-consuming.

In this section, we introduce the first stage of the \mval{} pipeline. We automatically inject structural variation into SAE text using linguistic perturbation rules that alter syntax and morphology but preserve semantics. In this way, perturbations preserve labels. Unlike many black-box translation approaches
\citep{krishna-etal-2020-reformulating,sun2022dialect}, label preservation will allow users to convert existing benchmarks directly into dialectal stress tests. Modular, independent perturbation functions give researchers the flexibility to isolate the effects of different features in different combinations. 

What distinguishes our work from other syntactic data augmentation methods \citep{wu2022oolong} is that our perturbations are grounded in formal language patterns. We operationalize the decades of linguistics research cataloged in the Electronic World Atlas of Varieties of English (eWAVE; \citeauthor{ewave} \citeyear{ewave}), a database with 235 features from 75 English varieties, as documented by 87 professional linguists in 175 peer-reviewed publications. eWAVE distinguishes dialects by their unique clusters of linguistic features and the relative \textit{pervasiveness} of each feature.\footnote{For example, the \textit{give passive} feature \#\ewave{153} is considered pervasive or obligatory in Colloquial Singapore English, while it is rarely observed in Philippine and Tristan da Cunha English, and it is never seen in any other dialect.}
We define a \textbf{dialect transformation} as a sequential application of perturbation rules. Decisions to perturb the text follow the eWAVE heuristic probabilities: 100\% for obligatory features; 60\% for features neither pervasive nor rare; 30\% for rare features; 0\% for features with no information or an attested absence.

\begin{figure}
    \centering
    \includegraphics[width=0.8\columnwidth]{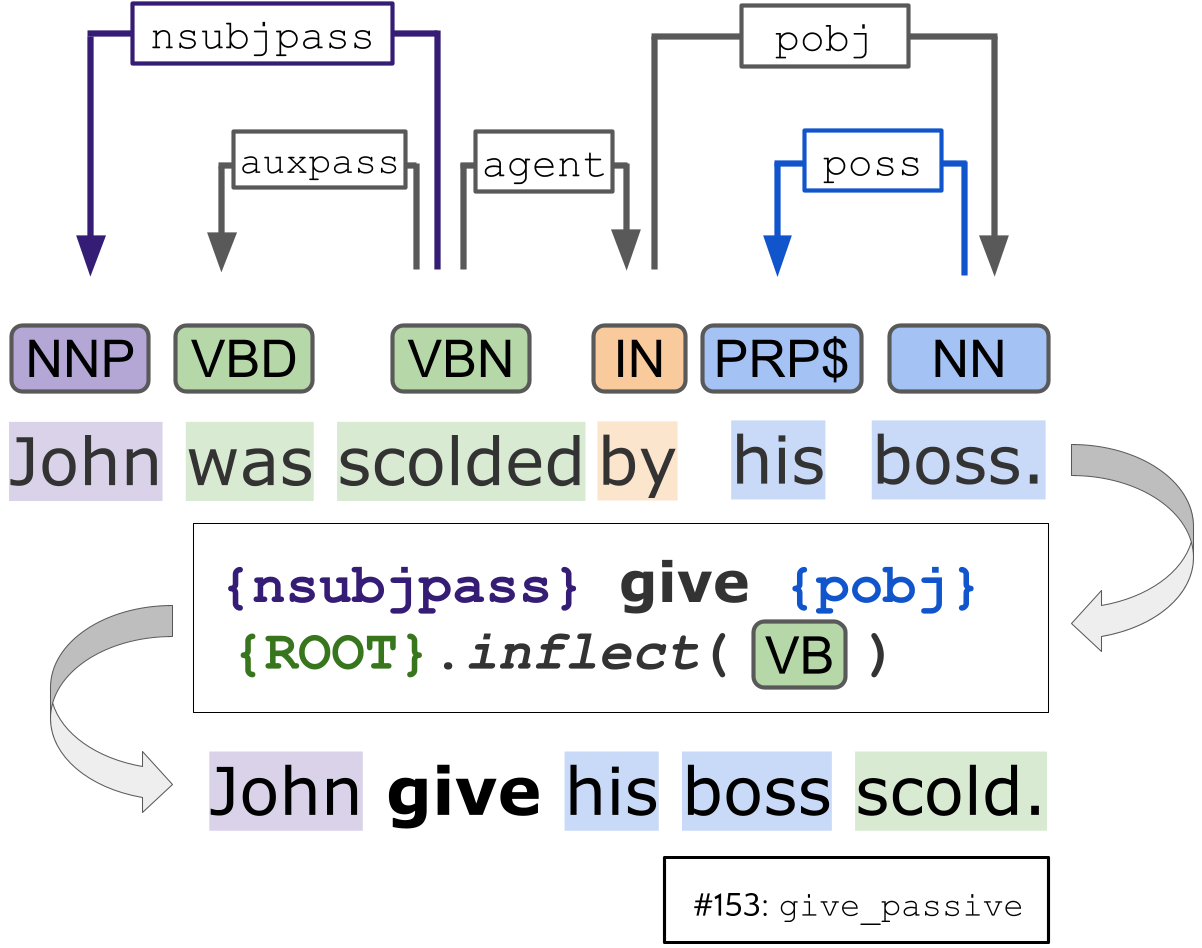}
    \caption{ The \texttt{give\_passive} [\#\ewave{153}] perturbation follows this procedure: (1) take the passive subject (\texttt{nsubjpass}); (2) insert the verb \textit{give}; (3) insert the object of a preposition that serves as the \texttt{agent} of the \texttt{ROOT}, (4) insert the \texttt{ROOT}, inflected with its base form.}
    \label{fig:ex_give_passive}
\end{figure}

For each rule, we condition the perturbation on morphosyntactic signals from POS tags, noun and verb inflection, and dependency relations using the \href{https://spacy.io/}{\texttt{spaCy 2.1.0}} \citep{honnibal2020spacy} and \href{https://pypi.org/project/inflect/}{\texttt{inflect 5.5.2}} libraries. For the \textit{give passive} pertubation above in Figure~\ref{fig:ex_give_passive}, we search for passive constructions with a past participle \texttt{ROOT} (\texttt{VBN}), an \texttt{nsubjpass} patient, and an agent. We construct the new phrase by inflecting the \texttt{ROOT} to its base (\texttt{VB}) form and moving it after the entire agentive noun phrase.

Following the eWAVE organizational scheme, we motivate and present our feature perturbations in 12 grammatical categories: (\hyperref[subsec:pronouns]{1}) \nameref{subsec:pronouns}, (\hyperref[subsec:nounphrases]{2}) \nameref{subsec:nounphrases}, (\hyperref[subsec:tenseaspect]{3}) \nameref{subsec:tenseaspect}, (\hyperref[subsec:mood]{4}) \nameref{subsec:mood}, (\hyperref[subsec:verbmorphology]{5}) \nameref{subsec:verbmorphology}, (\hyperref[subsec:negation]{6}) \nameref{subsec:negation}, (\hyperref[subsec:agreement]{7}) \nameref{subsec:agreement}, (\hyperref[subsec:relativization]{8}) \nameref{subsec:relativization}, (\hyperref[subsec:complementation]{9}) \nameref{subsec:complementation}, (\hyperref[subsec:adverbialsubordination]{10}) \nameref{subsec:adverbialsubordination}, (\hyperref[subsec:adverbsprepositions]{11}) \nameref{subsec:adverbsprepositions}, and finally (\hyperref[subsec:discoursewordorder]{12}) \nameref{subsec:discoursewordorder}. For a more detailed breakdown\hide{of every feature perturbation we implemented}, see Appendix~\ref{appdx:implementation_details}. 

\paragraph{Pronouns}
\label{subsec:pronouns}
\hide{are substitutes for noun phrases that typically involve nominal and temporal anaphora \citep{partee1984nominal}. Thus an understanding of pronouns is} are critical for tasks like machine translation and summarization, which depend on coreference resolution \citep{sukthanker2020anaphora}.\hide{, as well as current fairness research \citep{qian2022perturbation}.} Our pronoun perturbation rules account for linguistic structure and are not merely surface manipulations. For example, we condition on coreference for referential pronouns and on verb frames to identify benefactive datives. In total, we implement 39 of the 47 pronoun features from eWAVE.

\paragraph{Noun Phrases}
\label{subsec:nounphrases}
\hide{function as the arguments of verbs \citep{dryer2007noun}. Unsurprisingly, noun phrases} are the focus of fundamental NLP research in semantic role labeling and named entity recognition as well as downstream tasks like sentiment analysis, information extraction, summarization, and question answering \citep{gildea2002automatic}. \mval{} has 31 rules that operate on NP constituents. \hide{For example, \texttt{adj\_postfix} (feature \#\ewave{87}) moves attributive \texttt{AdjP} constituents \textit{after} the head noun they modify, as in the sentence ``I drive \textit{a car small and red}.''}

\paragraph{Tense and Aspect}
\label{subsec:tenseaspect}
are two grammatical properties that have to do with time. Together, these categories are known to significantly challenge machine translation \citep{matusov2019challenges,koehn2017six}.\hide{, among other applications.} With 26 rules, \mval{} introduces different kinds of inflections and auxiliary verbs to indicate when an action, event, or state occurred and how it extends over time.

\paragraph{Mood}
\label{subsec:mood}
\hide{is a grammatical category that has to do with \textit{modality}, or the relationship between the utterance and reality, possibility, desirability, or necessity \citep{pyatkin2021possible}. Modality} is important for applications in sentiment analysis and opinion mining, including the detection of biased language \citep{recasens2013linguistic} and framing strategies in political discourse \citep{king-morante-2020-must,demszky2019analyzing,ziems2021protect}. Misunderstandings of modality can also challenge NLU systems on tasks like natural language inference \citep{gong2018natural}. There are three modal perturbations in \mval{}.

\paragraph{Verb Morphology}
\label{subsec:verbmorphology}
is expected to affect model understanding of verb frame semantics \citep{baker1998berkeley}, which could impact performance on semantic role labeling, summarization, and machine translation, among other tasks. We implement 16 related perturbations that change verb suffixes, the forms of verb inflection, and the expression of semantic roles using specialized verbal phrases.

\paragraph{Negation}
\label{subsec:negation}
is covered by 16 eWAVE features, 14 of which are implemented in \mval{}. Problems with negation account for many of the failure cases in natural language inference \citep{hossain2020analysis} and sentiment analysis \citep{barnes2021improving}. Our perturbations introduce negative concord, invariant question tags, and new words for negation.

\paragraph{Agreement}
\label{subsec:agreement}
is a group of 11 rules which have to do with subject-verb agreement and the omission of copula and auxiliary \textit{be} in different environments. Examples include the invariant present tense in \textit{He speak English} (feature \#\ewave{170}), and the existential dummy word in \textit{It's some food in the fridge} (feature \#\ewave{173}). Nine of these 11 agreement features are attested in African American English (see \citeauthor{green2002african} \citeyear{green2002african}), which may be linked to the demonstrable performance disparities in AAVE dependency parsing \citep{blodgett2018twitter}, POS tagging \citep{jurgens-etal-2017-incorporating}, and NLU tasks \citep{ziems-etal-2022-value}.

\paragraph{Relativization}
\label{subsec:relativization}
is a class of perturbations that operates on relativizers\hide{---the group of \textit{wh}-pronouns, adverbs, and conjunctions used to }, which link relative clauses with their nouns. The purpose of a relative clause is to modify a noun phrase. It's an important construction for NLU because it can contain a presupposition \citep{joshi1976computation}. Our perturbation rules cover all 14 eWAVE features, operating both on individual relativizer words as well as sentence structure to move the relative clause and build correlative constructions, for example.

\paragraph{Complementation}
\label{subsec:complementation}
is a set of perturbations that turn dependent clauses into the subject or object of the sentence. \hide{A subordinating conjunction is a word that connects an independent clause with a dependent clause. A complementizer is a subordinating conjuction that, in many cases, turns the dependent clause into the subject or the object of the sentence. The dependent clause is called the complement.} Like relative clauses, \hide{verbal} complementation can contain presuppositions and implicatures \citep{potts2002lexical}, which are critical for natural language understanding. They can also convey a speaker's degree of certainty \citep{couso2015epistemic}, which correlates with biased language and framing strategies. We implement all 11 complementation features that are catalogued in eWAVE.

\paragraph{Adverbial Subordination}
\label{subsec:adverbialsubordination}
is a set of perturbations that operate on independent clauses with a ``conjunctive adverb.'' Adverbial conjunctions can express causality (\textit{therefore}), purpose (\textit{so that}), sequence (\textit{then}), contrast (\textit{however}), comparison (\textit{similarly}), and various forms of emphasis (\textit{indeed}). We implement all 5 eWAVE features in this class.

\paragraph{Adverbs and Prepositions}
\label{subsec:adverbsprepositions}
are represented by four rules, which can drop prepositions and replace adverbs with their adjectival forms.

\paragraph{Discourse and Word Order} has two sides:
\label{subsec:discoursewordorder}
two discourse features and 9 phrase-based perturbations that move entire constituents in a manner similar to \textit{constituency replacement} \citep{sutiono2022syntax}. These rules significantly alter the sentence structure, and in this way radically differ from prior token-level data augmentation techniques like synonym replacement \citep{wei-zou-2019-eda}. Phrasal movements include fronting and clefting, subject-auxiliary inversion, and a lack of inversion in questions. We also inject the word \textit{like} to indicate focus or quotation.

\section{Scope and Reliability of \mval{}}
\label{sec:validation_reliability}
\begin{figure*}
    \centering
    \includegraphics[width=0.94\textwidth]{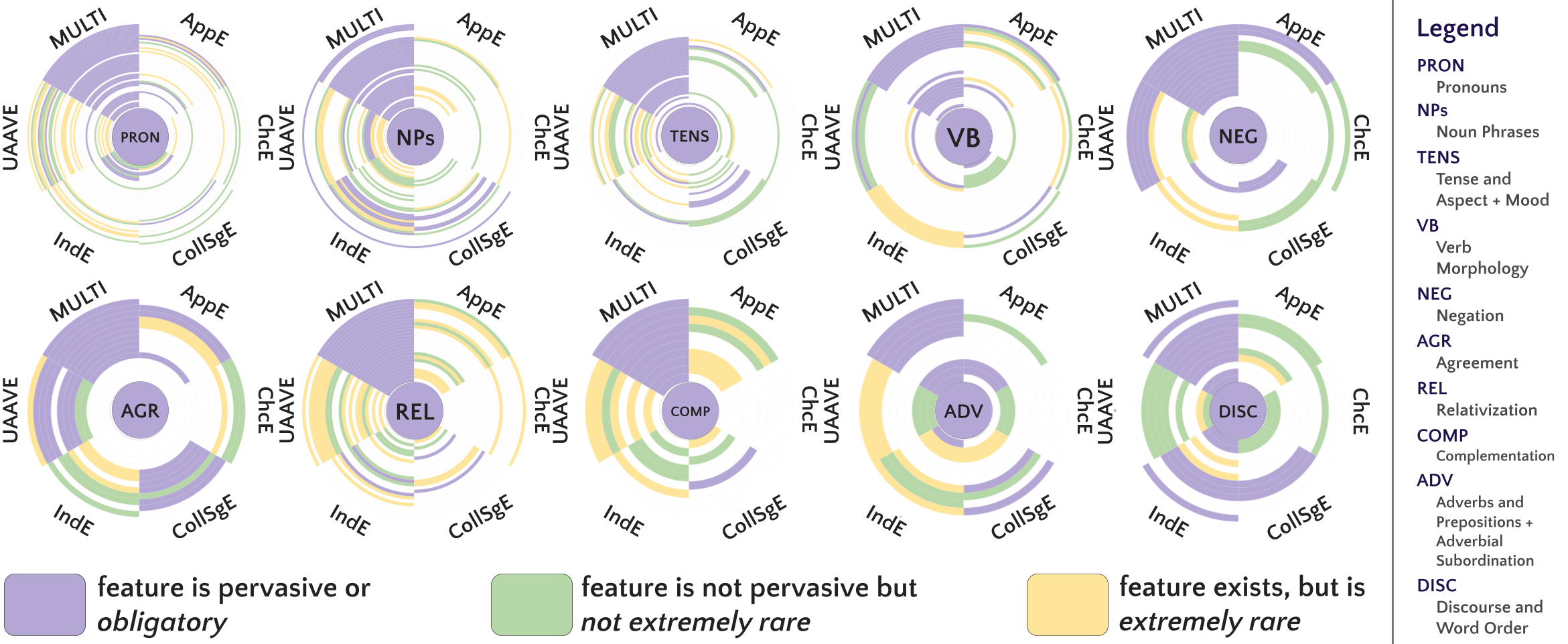}
    \caption{\textbf{A comparative distribution of the features in five dialects,} where a dialect is given by a slice of the wheel. Each wheel represents one of the feature groupings in \S\ref{sec:dialect_transformations} (\textbf{Tense and Aspect} + \textbf{Mood} are combined, as are \textbf{Adverbial Subordination} + \textbf{Adverbs and Prepositions}). The MULTI slice indicates which of the features are implemented in \mval{}. Rings indicate distinct features and are colored by pervasiveness in each dialect.}
    \label{fig:feat_wheels}
\end{figure*}

\subsection{Scope} 
\label{subsec:scope}
\mval{}'s scope is extensive.
Out of the 235 features documented in eWAVE, \mval{}
covers \numFeatures{}, spanning all \numDialects{} recorded English dialects. On average, the feature space for any given dialect is \averageDialectCoverage{} implemented, and no dialect is less than 80\% implemented (see Appendix~\ref{appdx:implementation_details}). \hide{The reason we did not cover 100\% of the eWAVE catalogue is that some features operate with information unavailable to us. For example, in SAE, aspect and mood may not be marked morphosyntactically; these features are outside the scope of current methods. Similarly, we are unable to inject distinct pronouns for groups of 2, 3, and 4+ people [\#\ewave{37}], as group size information may not be contained in the focus utterance.}

\vspace{-2pt}
\subsection{Recruiting Native Speakers for Validation} 

One key benefit of the \mval{} approach is our ongoing partnership with native speakers to confirm that our theoretically-inspired rules generate plausible and grammatical text. Here, we validate our transformation rules using the linguistic acceptability judgments of native speakers for 10 English dialects.\footnote{Chicano (29 annotators), Colloquial American (13), Indian (11), Appalachian (4), Aboriginal (4), North of England (3), Ozark (3), Southeast American Enclave (3), Urban African American (1), and Black South African English (1).} We recruit speakers from Amazon Mechanical Turk and screen them using a Dialect Assessment Survey.\footnote{\url{https://calebziems.com/resources/value/dialect-quiz.html}} This qualification survey ensures that each speaker's empirical language patterns
align with the literature on the dialect that they had self-reported. At each turn, the speaker considers a sentence in the target dialect and provides a binary grammaticality judgment about that sentence. Sentences come from published linguistics journals. The survey is efficient\footnote{The survey uses an average of 9 questions, but the survey length will depend upon the user's answers.} as it implements binary search, dynamically selecting the feature that most evenly partitions the space of candidate dialects. 

\subsection{Validating the \mval{} Pipeline}
\label{subsec:validation}

\begin{table}
\centering
\resizebox{\columnwidth}{!}{%
\fontsize{8.08pt}{8.08pt}\selectfont
\begin{tabular}{lc|lc|lc|lc}\toprule

\textsc{Feat.} & \textsc{Acc.} & \textsc{Feat.} & \textsc{Acc.}& \textsc{Feat.} & \textsc{Acc.} & \textsc{Feat.} & \textsc{Acc.} \\ \midrule
\ewave{10} & \valbest{\phantom{1}97.4} & \ewave{67} & \valbest{\phantom{1}99.1} & \ewave{128} & \valgood{\phantom{1}92.7} & \ewave{173} & \valgood{\phantom{1}87.5} \\
\ewave{39} & \valbest{\phantom{1}99.7} & \ewave{70} & \valgood{\phantom{1}92.9} & \ewave{130} & \valgood{\phantom{1}92.9} & \ewave{175} & \valbad{\phantom{1}83.3} \\
\ewave{40} & \valbest{\phantom{1}99.8} & \ewave{71} & \valbest{\phantom{1}98.8} & \ewave{132} & \valgood{\phantom{1}87.7} & \ewave{193} & \valgood{\phantom{1}88.7} \\
\ewave{42} & \valbest{\phantom{1}98.1} & \ewave{88} & \valbest{\phantom{1}99.4} & \ewave{133} & \valbest{\phantom{1}99.5} & \ewave{216} & \valbest{\phantom{1}99.7} \\
\ewave{43} & \valgood{\phantom{1}93.2} & \ewave{96} & \valbest{\phantom{1}95.5} & \ewave{154} & \valgood{\phantom{1}92.9} & \ewave{220} & \valbest{\phantom{1}99.4} \\
\ewave{49} & \valbest{\phantom{1}99.6} & \ewave{99} & \valgood{\phantom{1}94.7} & \ewave{155} & \valbad{\phantom{1}81.8} & \ewave{221} & \valgood{\phantom{1}86.7} \\
\ewave{56} & \valbest{\phantom{1}97.3} & \ewave{100} & \valbest{\phantom{1}99.9} & \ewave{165} & \valbest{\phantom{1}99.1} & \ewave{224} & \valbest{\phantom{1}99.5} \\
\ewave{60} & \valbest{\phantom{1}99.6} & \ewave{121} & \valgood{\phantom{1}91.8} & \ewave{170} & \valgood{\phantom{1}94.9} & \ewave{227} & \valgood{\phantom{1}91.2} \\
\ewave{63} & \valbest{\phantom{1}99.0} & \ewave{126} & \valgood{\phantom{1}92.3} & \ewave{172} & \valgood{\phantom{1}90.0} & \ewave{228} & \valbest{\phantom{1}99.8} \\ \midrule
\multicolumn{7}{l|}{\textsc{Feats.}} & \textsc{Acc.}\\ \midrule
\multicolumn{7}{m{8cm}|}{\ewave{3}, \ewave{9}, \ewave{11}, \ewave{14}, \ewave{15}, \ewave{16}, \ewave{26}, \ewave{29}, \ewave{33}, \ewave{34}, \ewave{41}, \ewave{45}, \ewave{47}, \ewave{55}, \ewave{57}, \ewave{58}, \ewave{59}, \ewave{61}, \ewave{62}, \ewave{64}, \ewave{66}, \ewave{77}, \ewave{78}, \ewave{79}, \ewave{80}, \ewave{81}, \ewave{86}, \ewave{101}, \ewave{106}, \ewave{117}, \ewave{119}, \ewave{123}, \ewave{131}, \ewave{134}, \ewave{145}, \ewave{146}, \ewave{149}, \ewave{159}, \ewave{174}, \ewave{179}, \ewave{191}, \ewave{194}, \ewave{198}, \ewave{203}, \ewave{204}, \ewave{205}, \ewave{206}, \ewave{207}, \ewave{208}, \ewave{209}, \ewave{214}, \ewave{223}, \ewave{226}, \ewave{232}, \ewave{235}} & \valbest{100.0} \\
\bottomrule 
\end{tabular}
}
\caption{ \textbf{Accuracy of 92 perturbation rules} according to majority vote with at least 5 unique sentence instances. Seventy four rules have \colorbox{valbest}{$\scriptstyle>95\%$} accuracy, while sixteen have accuracy in \colorbox{valgood}{$\scriptstyle[85, 95)$}, and only two are \colorbox{valbad}{$\scriptstyle<85\%$} accurate, demonstrating the reliability of our approach.
}
\label{tab:validation_stats}
\end{table}

To validate our perturbation rules, we use the task from \citet{ziems-etal-2022-value} in which each annotator is shown a pair of sentences: one in SAE, and the other as a dialect transformation: a copy of the first with perturbations corresponding to the target dialect. Annotators see only perturbations corresponding to their native dialect.  Annotators mark portions of sentence 1 that were perturbed incorrectly in sentence 2. The interface is shown in in Figure~\ref{fig:mturk_task} in the Appendix.

A group of 72 annotators evaluate a total of 19k sentence pairs, which were drawn from CoQA and other sources. We use CoQA sentences for our Gold Test Sets (\S\ref{subsec:gold}), and for added syntactic diversity, we pull sentences from three \texttt{nltk} corpora: Reuters \citep{russell2002reuters}, Sentiment Analysis \citep{pang2004sentimental} and Movie Reviews \citep{pang2005seeing}. Three annotators evaluate each transformation, marking any pre-highlighted spans where the transformation appeared ungrammatical. This gives us both transformation and perturbation-level evaluations. The majority vote determines the accuracy of the perturbation rule.\footnote{Accuracy reliably measures strong consensus in the quality of our approach and, unlike kappa scores, it will not suffer from the \textit{prevalence problem} \citep{eugenio2004kappa}.} Perturbation accuracies are given in Table~\ref{tab:validation_stats}. Since there are 55 rules with perfect accuracy, and all perturbation rules achieve above $81$\%, researchers can feel confident in the linguistic plausibility of the \mval{} transformation pipeline.

\subsection{Gold Test Sets} 
\label{subsec:gold}
While synthetic \mval{} transformations will be useful for identifying weak points in a model's performance, this does not ensure the model is ready for the real world. We urge practitioners to heavily test user-facing models with numerous in-domain tests. As a first step, we provide reliable gold standard CoQA datasets in Chicano English 
(ChcE) and Indian English (IndE). Out of 7,983 CoQA questions, our pipeline made changes to 1,726 ChcE questions (21.6\%) and 6,825 IndE questions (85.4\%). Human annotators considered only transformed questions and provided their own alternative phrasing for transformations they found ungrammatical. Alternatively, they could simply exclude the erroneous perturbations from the question. ChcE had a total transformation accuracy of 82.7\% while IndE had 66.1\%. The lower IndE accuracy is due to the higher density of features in this dialect. After rephrasing or removing errors, we were left with 1,498 dialect-transformed ChcE questions and 5,289 IndE questions. Together with any unperturbed questions, these gold questions constitute the gold test sets for evaluation in \S\ref{subsec:linking_natural_synth}.

\section{Using \mval{}}
\label{sec:using_mval}
With our feature rules written (\S\ref{sec:dialect_transformations}) and hand-validated by native speakers (\S\ref{sec:validation_reliability}), we can use \mval{} to create synthetic data for training dialect-robust models and also for stress testing leading systems on dialect benchmarks. We specifically provide synthetic data for five English dialects:
%\diyi{are these mainly within the United States? if so, make it explicit in our writing}: 
Appalachian (AppE), 
Chicano English (ChcE),
Indian English (IndE), 
Colloquial Singapore English (CollSgE),
and Urban African American English (UAAVE).
Three of these dialects are based in the US, where annotators were most abundant for validation, and two are outside the US.

To understand models' ability to transfer knowledge between dialects, we also consider models trained on dialect $A$ and evaluated on dialect $B$ for each dialectal pair $(A,B)$. We can further leverage the strengths of \mval{} as a multi-dialectal augmentation tool by training on a synthetic pseudo-dialect that contains the union of all feature options \textbf{(Multi)}. We hypothesize that models trained on multi-(pseudo)-dialectal data will benefit from robustness. \hide{to transformations across dialects.}
While the \mval{} approach could apply over any task with free-form text, we focus on three domains in particular: conversational question answering, semantic parsing, and machine translation. All three are user-facing tasks where language variation may hinder users' access to information, resources, and/or the global economy \citep{blasi2022systematic,faisal2021sd}.

\paragraph{Conversational Question Answering} (CoQA; \citeauthor{reddy2019coqa}\citeyear{reddy2019coqa}) is a reading comprehension benchmark with 127k question-answer pairs and 8k passages in seven different genres and domains. We use it because it is a challenging task where dialect-induced errors can compound. The primary challenge is that questions are conversational: they contain coreference and pragmatic relations to prior questions. To transform the publicly available training and development sets, we perturb only questions. This is a natural information-retrieval setting: the user submits queries in a low-resource dialect while the underlying corpus is in SAE.

\paragraph{Semantic Parsing} is the task of mapping natural language to formal language. This is a critical skill for dialogue systems, information retrieval, code generation, and other user-facing applications where dialect use is likely. We transform Spider \citep{yu2018spider}, a widely-used text-to-SQL benchmark. Again, we transform only the natural language query, leaving both the database tables and the SQL query unchanged to simulate interaction with a dialect user. Unlike the question answering setting where knowledge is encoded in free-text SAE passages, the knowledge and query language in Spider are encoded in formal tables and structured language, both of which are dialect-free. Consequently, any performance discrepancies here will be due to a mismatch between the models' training and testing data rather than a mismatch between the query dialect and that of the knowledge base. 

\paragraph{Machine Translation} is an interesting test case where challenges can arise from domain mismatch \citep{koehn2017six} due to dialect. We especially anticipate challenges with verb morphology (\S\ref{subsec:verbmorphology}), tense and aspect (\S\ref{subsec:tenseaspect}), and pronouns (\S\ref{subsec:pronouns}). We use a standard dataset, WMT19, and evaluate translation from each English Dialect to Chinese, German, Gujurati, and Russian. This simulates a user interacting with translation software using their native dialect.

\section{Cross-Dialectal Stress Testing}
\label{sec:benchmarking_experiments}

\begin{table}
\centering
\resizebox{1\columnwidth}{!}{%
\begin{tabular}{c|c|c|cc}
\toprule
\multicolumn{2}{c|}{Model}                   & \multicolumn{3}{c}{Test Dialect} \\ \midrule
Base                           & Train Set  & SAE & ChcE  & IndE                              \\ \midrule
                               & SAE       & 77.2 & 76.7 (-0.5\%) & 72.3 \drop{(-6.7\%)} \\
                                   & Multi   & 76.2 (-1.2\%)   & 76.1 (-1.4\%) & 75.0 \drop{\up{(-2.9\%)}}\\ \cmidrule{2-5}
\multirow{-4}{*}[-4pt]{\rotatebox{90}{BERT}}    & In-Dialect & 77.2 & 76.5 (-0.9\%) & 75.1 \drop{\up{(-2.7\%)}} \\ \midrule
                               & SAE       & 81.8 & 81.6 (-0.2\%) & 77.7 \drop{(-5.2\%)} \\
                               & Multi      & 80.6 \drop{(-1.5\%)} & 80.5 \drop{(-1.6\%)} & 79.7 \drop{\up{(-2.7\%)}}\\ \cmidrule{2-5}
\multirow{-4}{*}[-4pt]{\rotatebox{90}{\small RoBERTa}} & In-Dialect & 81.8 & 81.6 (-0.2\%) & 80.5 \drop{\up{(-1.6\%)}} \\ \bottomrule
\end{tabular}}
\caption{\textbf{Gold QA Evaluation:} F1 Metric on each gold development set of the CoQA benchmark. \drop{} and \up{} respectively indicate significantly ($P < 0.05$) worse performance than SAE$\mapsto$SAE and better performance than SAE$\mapsto$Dialect by a paired bootstrap test.
}
\label{tab:gold_performance}
\end{table}

\begin{table*}[h]
\centering
\resizebox{\textwidth}{!}{%
\fontsize{8.08pt}{8.08pt}\selectfont
\begin{tabular}{c|c|c|ccccc|cH}
\toprule
\multicolumn{2}{c|}{Model}                   & \multicolumn{7}{c}{Test Dialect}                                                                                                                                                                                                                             \\ \midrule
Base                           & Train Set  & SAE           & AppE                                 & ChcE                                 & CollSgE                               & IndE                                 & UAAVE                                & Average & Avg.                              \\ \midrule
                               & SAE        & 77.2    & 74.4 \drop{(-3.8\%)} & 76.6 (-0.7\%)                        & 61.5 \drop{(-25.4\%)} & 70.8 \drop{(-9\%)}   & 71.2 \drop{(-8.4\%)} & 71.9 (-7.3\%) & 74.0 (-4.2\%) \\
                               & AppE       & 76.3 (-1.1\%) & 76.4 \up{(-1\%)}                          & 76.1 (-1.4\%)                        & 64.7 \up{\drop{(-19.3\%)}} & 72.8 \up{\drop{(-6\%)}}   & 73.2 \up{\drop{(-5.4\%)}} & 73.3 (-5.3\%) & 75.0 (-2.9\%)\\
                               & ChcE       & 76.8 (-0.5\%) & 74.7 \drop{(-3.3\%)} & 76.5 (-0.8\%) & 63.6 \up{\drop{(-21.3\%)}} & 71.6 \drop{(-7.8\%)} & 71.4 \drop{(-8.1\%)} & 72.4 (-6.5\%) & 79.3 (-3.2\%)\\
\iftrue                              & CollSgE    & 75.7 \drop{(-1.9\%)} & 74.1 \drop{(-4.2\%)} & 75.5 \drop{(-2.2\%)} & 74.7 \up{\drop{(-3.3\%)}}                         & 73.6 \up{\drop{(-4.8\%)}}                        & 73.4 (-5.1\%) & 74.5 (-3.6\%)                        \\ \fi
                               & IndE       & 76.0 (-1.5\%) & 75.4 \drop{(-2.4\%)}                        & 75.7 \drop{(-2\%)}                          & 63.2 \up{\drop{(-22\%)}}   & 75.1 \up{\drop{(-2.7\%)}}                        & 74.1 \up{\drop{(-4.1\%)}}                        & 73.3 (-5.3\%) & 75.3 (-2.5\%)\\
                               & UAAVE      & 76.1 (-1.4\%) & 75.6 \up{\drop{(-2\%)}}                          & 76.0 \drop{(-1.5\%)}                        & 64.6 \up{\drop{(-19.5\%)}} & 74.5 \up{\drop{(-3.6\%)}}                        & 75.3 \up{\drop{(-2.5\%)}}                        & 73.7 (-4.7\%) & 75.5 (-2.2\%) \\
                               & Multi      & 76.2 (-1.2\%) & 75.6 \up{\drop{(-2\%)}}                          & 76.1 (-1.3\%)                        & 73.7 \up{\drop{(-4.7\%)}}                         & 74.9 \up{\drop{(-3.1\%)}}                        & 75.1 \up{\drop{(-2.7\%)}}                        & 75.3 (-2.5\%) & 75.6 (-2.1\%)                       \\ \cmidrule{2-10}
\multirow{-8}{*}[5pt]{\rotatebox{90}{BERT Base}}    & In-Dialect & 77.2    & 76.4 \up{(-1\%)}                          & 76.5 (-0.8\%)                        & 74.7 \up{\drop{(-3.3\%)}}                         & 75.1 \up{\drop{(-2.7\%)}}                        & 75.3 \up{\drop{(-2.5\%)}}                        & 75.9 (-1.7\%) & 76.1 (-1.4\%)                       \\ \midrule
                               & SAE        & 81.8    & 79.1 \drop{(-3.4\%)} & {81.5 (-0.3\%)} & 68.8 \drop{(-18.9\%)} & 76.1 \drop{(-7.5\%)} & 76.6 \drop{(-6.7\%)} & 77.3 (-5.8\%) & 79.0 (-3.5\%)\\
                               & AppE       & 82.0 (0.3\%)  & 81.8\up{}                           & 81.8                           & 71.2 \up{\drop{(-14.9\%)}} & 79.0 \up{\drop{(-3.5\%)}}                        & 79.6 \up{\drop{(-2.8\%)}}                        & 79.2 (-3.2\%) & 80.8  (-1.2\%)\\
                               & ChcE       & 81.7 (-0.1\%) & 79.3 \drop{(-3.1\%)} & 81.5 (-0.4\%)                        & 68.8 \drop{(-18.9\%)} & 76.5 \drop{(-7\%)}   & 77.3 \drop{(-5.9\%)} & 77.5 (-5.5\%) & 79.3 (-3.2\%) \\
                              \iftrue   & CollSgE    & 81.5 (-0.4\%) & 80.1 \drop {(-2.2\%)} & 81.2 (-0.7\%)                        & 80.2 \up{\drop{(-2\%)}} & 79.4 \up{\drop{(-3\%)}} & 78.7 \up{\drop{(-3.9\%)}} & 80.2 (-2\%)                          \\ \fi
                               & IndE       & 81.1 (-0.8\%) & 80.5 \up{\drop{(-1.5\%)}}                        & 80.9 (-1.1\%)                        & 67.2 \drop{(-21.7\%)} & 80.3 \up{\drop{(-1.9\%)}}                        & 79.2 \up{\drop{(-3.3\%)}} & 78.2 (-4.6\%) & 80.4 (-1.7\%)\\
                               & UAAVE      & 81.6 (-0.2\%) & 81.1 \up{(-0.9\%)}                        & 81.5 (-0.3\%)                        & 69.2 \drop {(-18.2\%)} & 79.6 \up{\drop{(-2.7\%)}}                        & \up{81.1 (-0.9\%)}                        & 79.0 (-3.5\%) & 81.0 (-1\%)\\
                               & Multi      & 80.6 \drop{(-1.5\%)} & 80.4 \up{\drop{(-1.7\%)}}                        & 80.5 \drop{(-1.6\%)}                        & 78.5 \up{\drop{(-4.2\%)}}  & 79.7 \up{\drop{(-2.7\%)}}                        & 80.0 \up{\drop{(-2.2\%)}}                        & 80.0 (-2.3\%) & 80.2 (-1.9\%)                        \\ \cmidrule{2-10}
\multirow[c]{-8}{*}[5pt]{\rotatebox{90}{RoBERTa Base}} & In-Dialect & 81.8    & 81.8\up{}                           & 81.5 (-0.4\%)                        & 80.2 \up{\drop{(-2\%)}}                           & 80.3 \up{\drop{(-1.9\%)}}                        & 81.1 \up{(-0.9\%)}                        & 81.1 (-0.9\%) & 81.3 (-0.6\%)                       \\ \bottomrule
\end{tabular}
}
\caption{\textbf{Dialect QA Stress Test:} F1 Metric on each VALUE-transformed development set of the CoQA benchmark. \drop{} and \up{} indicate significantly ($P < 0.05$) worse performance than SAE$\mapsto$SAE and better performance than SAE$\mapsto$Dialect by a paired bootstrap test.
}
\label{tab:performance}
\end{table*}

\begin{table*}[]
\resizebox{\textwidth}{!}{%
\fontsize{8.08pt}{8.08pt}\selectfont
\begin{tabular}{c|c|c|ccccc|c}
\toprule
                                \multicolumn{2}{c|}{Evaluation}     & \multicolumn{7}{c}{Input Dialect}                                                                             \\ \midrule
                         Model & Metric & SAE & AppE         & ChcE         & CollSgE       & IndE         & UAAVE        & Avg.         \\\midrule
                         \multirow{2}{*}{BART-base}  & Exact Match ACC      & 49.3    &  45.2 \drop{(-8.3\%)} & 48.5 \drop{(-1.6\%)} & 41.9 \drop{(-15.0\%)} & 40.5 \drop{(-17.8\%)} & 45.0 \drop{(-8.7\%)}  & 45.1 (-8.5\%) \\
                            & Execution ACC        & 51.0    & 47.3 \drop{(-7.3\%)} & 50.3 (-1.4\%) & 44.1 \drop{(-13.5\%)} & 42.3 \drop{(-17.1\%)} & 46.1 \drop{(-9.6\%)}  & 46.9 (-8.0\%) \\ \midrule
\multirow{2}{*}{BART-large} & Exact Match ACC      & 67.9    & 63.6 \drop{(-6.3\%)} & 65.5 \drop{(-3.5\%)} & 60.3 \drop{(-11.2\%)} & 61.2 \drop{(-9.9\%)}  & 62.3 \drop{(-8.2\%)}  & 63.5 (-6.5\%) \\
                            & Execution ACC        & 70.5    & 65.2 \drop{(-7.5\%)} & 68.2 \drop{(-3.3\%)} & 63.0 \drop{(-10.6\%)} & 62.8 \drop{(-10.9\%)} & 64.5 \drop{(-8.5\%)}  & 65.4 (-7.2\%) \\\midrule
\multirow{2}{*}{T5-base}    & Exact Match ACC      & 58.7    & 54.3 \drop{(-7.5\%)} & 57.4 \drop{(-2.2\%)} & 50.0 \drop{(-14.8\%)} & 49.1 \drop{(-16.4\%)} & 53.1 \drop{(-9.5\%)}  & 53.8 (-8.3\%) \\
                            & Execution ACC        & 59.8    & 56.0 \drop{(-6.4\%)} & 58.5 \drop{(-2.2\%)} & 51.6 \drop{(-13.7\%)} & 51.3 \drop{(-14.2\%)} & 54.6 \drop{(-8.7\%)}  & 55.3 (-7.5\%) \\\midrule
\multirow{2}{*}{T5-3b}      & Exact Match ACC      & 71.7    & 65.3 \drop{(-8.9\%)} & 69.7 \drop{(-2.8\%)} & 60.7 \drop{(-15.3\%)} & 62.9 \drop{(-12.3\%)} & 68.5 \drop{(-4.5\%)}  & 66.5 (-7.3\%) \\
                            & Execution ACC        & 75.6    & 69.3 \drop{(-8.3\%)} & 73.4 \drop{(-2.9\%)} & 64.9 \drop{(-14.2\%)} & 66.5 \drop{(-12.0\%)} & 66.9 \drop{(-11.5\%)} & 69.4 (-8.2\%)\\
\bottomrule
\end{tabular}
}
\caption{\textbf{Dialect SPIDER Stress Test:} Evaluation on each VALUE-transformed evaluation set of the SPIDER benchmark. We finetune BART and T5 on SPIDER and evaluate for both Exact Match and Execution accuracy. \drop{} indicates a significant performance drop ($P < 0.05$) compared to SAE performance by a bootstrap test.}
\label{tab:sqlperformance}
\end{table*}

\begin{table*}[t]
\resizebox{2\columnwidth}{!}{%
\begin{tabular}{c|c|c|ccccc|c}\toprule
                                     \multicolumn{2}{c|}{Evaluation}                 & \multicolumn{7}{c}{Source Dialect}                                                                             \\ \midrule
\# Param.                                & Target & SAE  & AppE            & ChcE           & CollSgE         & IndE            & UAAVE          & Avg.        \\ \midrule
\multirow{4}{*}{615M} & Chinese         & 22.5	&	21.2 \drop{(-6.1\%)} & 21.7 \drop{(-3.6\%)}	& 17.0 \drop{(-24.5\%)} & 18.7 \drop{(-16.8\%)} & 19.8 \drop{(-12.3\%)} & 20.1 (-10.6\%) \\
                                     & German          & 39.6 & 34.3 \drop{(-13.41\%)} & 37.8 \drop{(-4.65\%)} & 22.3 \drop{(-43.60\%)} & 26.8 \drop{(-32.32\%)} & 30.5 \drop{(-23.1\%)} & 31.9 (-19.5\%) \\
                                     & Gujurati  & 21.7	& 18.6 \drop{(-14.5\%)}& 20.4 \drop{(-6.2\%)}& 13.4 \drop{(-38.4\%)} & 16.6 \drop{(-23.4\%)} & 17.2 \drop{(-20.7\%)} & 18.0 (-17.2\%)\\
                                     & Russian & 27.8 & 24.6 \drop{(-11.4\%)} & 26.7 \drop{(-4.0\%)} & 17.2 \drop{(-38.1\%)} & 20.8 \drop{(-25.4\%)} & 21.7 \drop{(-22.1\%)} & 23.1 (-16.8\%) \\ \midrule
\multirow{4}{*}{1.3B}           & Chinese         & 23.2 & 21.5 \drop{(-7.4\%)} & 22.5 (-3.3\%) & 17.8 \drop{(-23.5\%)} & 19.4 \drop{(-16.6\%)}	& 19.8 \drop{(-15.0\%)} &	20.7 (-11.0\%)\\
                                     & German          & 42.6 & 37.5 \drop{(-11.9\%)} & 40.6 \drop{(-4.6\%)} & 25.3 \drop{(-40.6\%)} & 29.4 \drop{(-31.0\%)} & 34.2 \drop{(-19.7\%)} & 34.9 (-18.0\%) \\
                                     & Gujurati        & 24.0 & 20.7 \drop{(-13.8\%)} & 22.9 \drop{(-4.5\%)} & 15.5 \drop{(-35.4\%)} & 18.5 \drop{(-22.8\%)} & 19.7 \drop{(-17.8\%)} & 20.2 (-15.7\%) \\
                                     & Russian         & 31.7 & 28.5 \drop{(-10.1\%)} & 30.3 (-4.4\%)	& 20.3 \drop{(-36.0\%)} & 24.5 \drop{(-22.6\%)} & 25.3 \drop{(-20.2\%)} & 26.7 (-15.5\%) \\ \bottomrule
\end{tabular}}
\caption{\textbf{Dialect Translation Stress Test:} SacreBLEU Score~\citep{sacrebleu} on each VALUE-transformed validation set of the WMT19 benchmark at 2 distilled scales of the NLLB Translation model~\citep{nllb}. \drop{} indicates a significant performance drop ($P < 0.05$) compared to SAE performance by a bootstrap test.}\label{tab:trans_perf}
\end{table*}

Here we benchmark current models on dialect variants of the three tasks in \S\ref{sec:using_mval}. For each dataset, we use fixed hyperparameters without early stopping and report all performances on dialect variants of the \textit{evaluation} data, since public test sets are not available for the original datasets. We use the base versions of BERT~\citep{devlin-etal-2019-bert} and RoBERTa~\citep{liu2019roberta} on dialect variants of the CoQA task, following the Rationale Tagging Multi-Task setup of \citet{method}. For SPIDER, we evaluate BART and T5, since both are near the state of the art in semantic parsing \citep{UnifiedSKG}. For Translation, we evaluate the NLLB Translation Model at two distilled scales: 615M and 1.3B \citep{nllb}. We report hyperparameters and further motivation for model selection in Appendix \ref{appdx:models_hypers}.

\subsection{Linking Natural and Synthetic Data}
\label{subsec:linking_natural_synth}
While natural data is the gold standard, it is difficult to scale to the number of dialects and tasks we can cover with synthetic data. Thus our broad evaluations are synthetic stress tests. Importantly, we first demonstrate the critical relationship between the gold and synthetic transformations using the gold evaluation sets from \S\ref{subsec:gold} and the synthetic training data from \S\ref{sec:using_mval}. Table \ref{tab:gold_performance} shows the gold standard CoQA results, which should be compared to the synthetic CoQA results in Table \ref{tab:performance}.

The synthetic stress test results match the gold performance for Chicano English with only small deviations. The Indian English stress tests slightly overestimate the performance drop of an SAE model on Indian English (70.8\% synthetic vs. 72.3\% natural IndE with BERT; 76.1\% vs. 77.7\% with RoBERTa). This is expected, as the synthetic feature density may be higher than some annotators naturally use. Synthetic results are a lower bound on performance for a target dialect. For all treatments, the stress tests are directionally correct: treatments that improve performance on the stress test also improve results on the gold data. 

Combined with speaker validation of the patterns themselves in \S\ref{subsec:validation}, this shows that \mval{} can be used to reliably measure the effects of modeling choices on dialectal performance.
\subsection{Synthetic Stress Tests}
We run 3 stress tests to understand worst-case performances on dialect-shifted data across a suite of models and tasks. Evaluation reveals large and statistically significant performance gaps across each task and across all dialects. This highlights, for the first time, the pervasiveness of English dialect disparity beyond any single dialect.

\paragraph{CoQA + Data Augmentation} results are shown in Table~\ref{tab:performance}. As predicted in \S\ref{subsec:linking_natural_synth}, Chicano English (ChcE) does not produce a significant drop in performance (-0.7\% BERT; -0.3\% RoBERTa) since few of its pervasive features are distinct from SAE (the Manhattan distance between feature vectors for ChcE and Colloquial American English is 0.14, or only half the distance as between CollAmE and CollSgE, IndE, and UAAVE.) On the other hand, Singapore English, which is distant from SAE and therefore has many obligatory features, leads to the largest drop (-25.4\% BERT; -18.9\% RoBERTa). Appalachian, Indian, and Urban African American English each induce significant but smaller RoBERTa performance drops of \mbox{-3.4\%}, \mbox{-7.5\%}, and \mbox{-6.7\%} respectively. 

The data augmentation technique described in \S\ref{sec:using_mval} successfully closes the dialectal performance gap. Across every dialect but Chicano English, we find that we can improve results by training on data that was transformed to the target dialect. Compared to standard RoBERTa, the RoBERTA model trained on \textbf{Multi}-dialectal data improves average cross-dialectal performance by 2.7 points. However, multi-dialectal training causes a drop of 1.2 points on SAE, reminiscent of interference in multilingual models~\citep{wang2019characterizing, wang2020negative}.\smallskip

We performed a \textbf{Qualitative Error Analysis} on 30 errors for each transformed dialect. In each error, models trained on SAE flipped from a correct answer in SAE to an incorrect answer in one of the dialect-transformed COQA sets. Fully validated perturbations in tense, inflection, plural marking, phrasal order, and the deletion of pragmatically-recoverable pronouns, prepositions, and auxiliaries all lead to significant errors. As expected, these errors can cascade down the conversation, leading to model failure on later \textit{unperturbed} questions as well. In some cases, erroneous answers still belong to the correct class, like flipping from \textit{yes} to \textit{no} in the presence of \textit{negative concord}. Suprisingly, transformations also frequently cause the model to respond with an erroneous \textit{class}, like giving a noun phrase or prepositional phrase to a yes/no question under perturbations like \textit{clefting} and the omission of auxiliary \textit{did}, \textit{is}, and \textit{wh}-words.

Our analysis also suggests that the noticeably larger drop in performance on Singapore English might be largely due to the higher density of two perturbation types: preposition omissions (feature \#\ewave{198}), and the \textit{one relativizer} (feature \#\ewave{216}). Future work can use perturbation analyses \citep{ziems-etal-2022-value} to quantitatively measure these sources of error.

\paragraph{Semantic Parsing} Table~\ref{tab:sqlperformance} shows that SAE models significantly underperform on all dialectal stress tests, both in terms of Exact Match Accuracy and Execution Accuracy. For both BART and T5, the largest performance gaps appear when we test on the two non-American dialects, CollSgE and IndE (-15.3\% and -12.3\% exact match accuracy for T5-3b). The semantic parsing performance gaps here are as large as those in conversational question answering. This supports our claim that the discrepancies are caused by model mismatch, rather than solely a mismatch between the dialect of the question and that of the knowledge base.

\paragraph{Machine Translation}
stress test results are shown in Table ~\ref{tab:trans_perf}. Except for ChcE, performance drops significantly across all dialects for each language.
Interestingly, the size of the average dialectal performance gap is higher when the target language is structurally \textit{more similar} to English: the largest average drop is from English$\mapsto$German~(\mbox{-19.5\%} on 615M; \mbox{-18.0\%} on 1.3B) and the smallest average drop is from English$\mapsto$Chinese~(-10.6\% on 615M; -11.0\% on 1.3B). This result cannot be explained simply as a reflection of the model's SAE translation performance. If it were, we might expect a smaller performance gap for Gujurati, a low-resource Indo-European language, since it has low SAE translation performance (21.7 SacreBLEU on 615M), but in fact, English$\mapsto$Gujurati has the second \textit{largest} dialectal translation performance gap~(\mbox{-17.2\%} on 615M; \mbox{-15.7\%} on 1.3B). Our explanation is that Gujurati has syntax that is more similar to English.

Despite both the 1.3B and 615M NLLB models being distilled from the same larger model, we see that the dialectal gap is smaller for German, Gujurati, and Russian. This suggests that model compression may affect low-resource dialects more heavily than SAE, similar to multi-lingual findings for low-resource languages~\citep{ahia-etal-2021-low-resource}. 

\section{Conclusion}
In this work, we introduced \mval{} -- a dialect robustness evaluation framework that is interpretable, flexible, scalable, responsible, and generalizable. The rule-based methods form a transparent syntactic translation system that can flexibly adjust to the shifting feature space of living dialects. Additionally, the transformation rules are reliably sourced from over a decade of linguistics literature and vetted by native speakers. After showing that these transformations predict human-translated dialect benchmark performance, we used them to build dialect benchmarks and training data at scale, without the need for additional annotation efforts. By training and evaluating in a cross-dialectal manner, we demonstrated how \mval{} can be used for more generalizable findings about model performance and dialect transferability. \mval{} can facilitate a wide range of NLP tasks and applications, such as measuring the relationships between dialect similarity and generalization performance, the scaling laws of dialect disparity, as well as inspiring algorithms on better dialect transfer. Overall, we anticipate that \mval{} will continue to support the development of more fair and equitable language technologies.

\section{Limitations}
\label{sec:limitations}
{Lexical variation} is not our focus because it is not well-described by systematic, scalable, and generalizable rules. One can derive lexical distributions from data, but many low-resource dialects lack corpora on which to base these insights. This is an important problem for future research.

Multi-VALUE's strength is its extensive coverage of English morphosyntacic patterns that have been documented in eWAVE by over 80 linguists. Such comprehensive resources are not available for other languages, but we encourage continued collaborations between computer scientists and linguists to build these resources for dialect-robust NLP systems across languages. As it stands, the current iteration of Multi-VALUE provides global value by serving a global contact language, English, and its 50 most documented varieties.

Despite the scope and precision of eWAVE for English, its catalog ultimately derives from linguists' oral interviews with native speakers, and here we can identify some additional limitations. First, the orthographic conventions that linguists use to encode spoken dialect may not always align with the speakers' own writing conventions and usage. Second, our approach can only cover the variation that linguists observe frequently enough to document, and in canonical forms in which they are documented. This means we may not fully capture variation within each feature. 

Finally, dialects should not be treated like deterministic speech patterns, but rather like a range of grammatical options or switches that may be turned on and off and adjusted for frequency in various social and personal contexts. Dialects do not always fit into nicely prescribed categories.

\section{Ethical Considerations}
This work makes use of human subjects for annotation. All procedures were subject to ethical review and were approved by the authors' institution. Consent was gathered in accordance with the authors' institution guidelines and annotators had access to a data use statement when giving consent.

The purpose of Multi-VALUE is to provide tools which enable researchers and practitioners to understand and mitigate dialectal bias in their models. We will release these tools responsibly, ensuring that users sign
a Data Use Agreement that forbids the use of \mval{} for deception, impersonation, mockery, discrimination, hate speech, targeted harassment and cultural appropriation.

In the agreement, researchers and practitioners will also acknowledge the Limitations of this work (\S\ref{sec:limitations}), that \mval{} may not fully or accurately represent the natural usage patterns of all sub-communities of speakers. \mval{} is designed to be easily updatable and configurable such that it can be extended by and for specific sub-communities and updated as dialects evolve over time.

\section*{Acknowledgements}  We are thankful to the members of SALT Lab for their helpful feedback on the draft. Caleb Ziems is supported by the NSF Graduate Research Fellowship under Grant No. DGE-2039655.
Part of this work was funded by an Amazon Faculty Research Award on Alexa Fairness in AI to DY.

\bibliographystyle{acl_natbib}
\bibliography{refs}

\appendix

\section{Implementation Details}
\label{appdx:implementation_details}

In Table~\ref{tab:dialect_implementations}, we give summary statistics for the number of features implemented for each of the \numDialects{} focus dialects, and the number of such features which were validated by native speakers. On average, the feature space for any given dialect is \averageDialectCoverage{} implemented, and no dialect is less than 80\% implemented. The reason we did not cover 100\% of the eWAVE catalogue is that some features operate with information unavailable to us. For example, in SAE, aspect and mood may not be marked morphosyntactically; these features are outside the scope of current methods. Similarly, we are unable to inject distinct pronouns for groups of 2, 3, and 4+ people [\#\ewave{37}], as group size information may not be contained in the focus utterance.

In Tables~\ref{tab:pronouns}-\ref{tab:discoursewordorder}, we detail our \mval{} implementations with an enumeration of our implemented dialects and features and examples of each. In the \textsc{Val Acc.} column we give the validation accuracy (\S\ref{subsec:validation}) as well as tags \gold{\textbf{ChcE}} or \gold{\textbf{IndE}} to indicate if the feature appears in the gold Chicano or Indian English CoQA dataset respectively.

\subsection{Pronouns}
There are 47 pronoun features in eWAVE, and we cover 39 of them (83\%). While simple regular expressions can cover some pronoun mappings, this is not always possible since English maps the same surface forms to different grammatical roles.\footnote{For example, \textit{her} is both the accusative in ``give it to her`` and the noun modifier in ``her cart,'' while the masculine pronouns in ``give it to him'' and ``his cart'' differ. This problem was observed but not solved in the rule-based perturbation augmentation of \citet{qian2022perturbation}.} We overcome this problem by conditioning rules on pronouns' syntactic roles. We also condition on coreference for referential pronouns [\ewave{29}], and on verb frames to identify benefactive datives [\ewave{9}]. Furthermore, we swap the morphology of possession [\ewave{20}], change reflexive marking [\ewave{11}-\ewave{16}], swap animate pronouns for inanimate objects [\ewave{1}-\ewave{2}], and include additional elements like reduplication [\ewave{40}]. In summary, our pronoun perturbation rules account for linguistic structure and are not merely surface manipulations.

\subsection{Noun Phrases}
Among our 31 noun phrase perturbations, we regularize or modify plural morphology [\ewave{49}] and comparison strategies [\ewave{80}], to drop or modify articles [\ewave{60}], construct phrases for possession [\ewave{75}], and adjust the tree adjoining order to create adjective postfixes [\ewave{87}].

\subsection{Tense and Aspect} 
Tense and aspect perturbations include alternative inflections and auxiliaries to mark tense [\ewave{117}],
including immediate vs. distant future [\ewave{119}],
as well as perfect aspect [\ewave{99}].

\subsection{Mood}
\mval{} includes perturbations that inject double modals [\ewave{121}] and quasi-modals [\ewave{126}], change verb inflections under modal scope [\ewave{123}], and introduce auxiliaries to mark the sequential or irrealis mood [\ewave{106}].

\subsection{Verb Morphology}
Verb morphology features include levelling certain finite and non-finite verb forms [\ewave{130}] adding suffixes for transitive verbs [\ewave{143}], and building \textit{serial verb phrases} \citep{tallerman2019understanding} to mark passive constructions [\ewave{153}], indirect objects [\ewave{148}], or the movement of direct objects [\ewave{150}].

\subsection{Negation}
\mval{} includes rules for building phrases with negative concord [\ewave{154}], and forms of negation with the negation words \textit{never}, \textit{no}, \textit{not}, \textit{no more} or \textit{ain't}, as well as special invariant tags for questions [\ewave{166}].

\subsection{Agreement}
We implement the invariant present tense [\ewave{170}], as well as the existential dummy \textit{it} [\ewave{173}].

\subsection{Relativization}
These perturbations modify the form of the relativizer [\ewave{186}-\ewave{190}], as well as drop [\ewave{193}] or introduce new shadow pronouns [\ewave{194}], such as double relativizers [\ewave{191}] and phrasal forms [\ewave{192}]. Our perturbations also operate on the sentence structure by forming correlative constructions [\ewave{196}], deleting stranded prepositions [\ewave{198}], and moving the relative clause before the head noun [\ewave{199}].

\subsection{Complementation}
These perturbations can change the form of the complementizer [\ewave{200}, \ewave{201}], delete [\ewave{208}, \ewave{209}] or introduce additional complementizer words [\ewave{203}, \ewave{204}], build existential constructions from complementizer phrases [\ewave{205}, \ewave{206}], and modify the verb in the non-finite clause complement [\ewave{210}].

\subsection{Adverbial Subordination}
Our perturbation rules introduce clause-final conjunctions [\ewave{211}, \ewave{212}] and double conjuctions [\ewave{214}, \ewave{215}], and remove the adverb in verb-chaining constructions [\ewave{213}], which together represent the five adverbial subordination features in eWAVE.

\subsection{Adverbial Prepositions}
In this section, we drop prepositions [\ewave{216}] and replace adverbs with their adjectival forms [\ewave{220}, \ewave{221}]. We also include the word \textit{too} as a qualifier [\ewave{222}].

\subsection{Discourse and Word Order}
In discourse, we insert the word \textit{like} as a focus [\ewave{234}] or quotation marker [\ewave{235}]. Our phrase-based perturbations include fronting and clefting [\ewave{223}, \ewave{224}], subject–auxiliary inversion in both negation phrases [\ewave{226}] and indirect questions [\ewave{227}], and a lack of inversion in certain questions [\ewave{228}, \ewave{229}]. These rules significantly alter the sentence structure, and in this way radically differ from prior token-level data augmentation techniques like synonym replacement \citep{wei-zou-2019-eda}. Our approach here is most similar to \textit{constituency replacement} \citep{sutiono2022syntax}.

\section{Models \& Hyperparameters}

\label{appdx:models_hypers}
\paragraph{CoQA} We use the base versions of BERT~\citep{devlin-etal-2019-bert} and RoBERTa~\citep{liu2019roberta} on dialect variants of the CoQA task, following the Rationale Tagging Multi-Task setup of \citet{method} to adapt these models to the CoQA setup which includes \textit{Yes, No,} and \textit{Unknown} responses in addition to extractive answers. Each model was trained on an Nvidia GeForce RTX 2080 Ti for approximately 6 hours. For each model and dialect, we fine-tune using AdamW~\citep{adamW} for 2 epochs with a batch size of 16 and a learning rate $3e-5$. 

\paragraph{Semantic Parsing.} Following \citet{UnifiedSKG}, for T5-base we adopted the AdamW optimizer, while Adafactor was used for T5-3B and the two BART models. We used NVIDIA A100 to train these models with T5-3b, BART-large, T5-base, and BART-base using 8 GPUs for 52 hours, 4 GPUs for 32 hours, 4 GPUs for 4 hours, 4 GPU for 13 hours respectively. We set the learning rate at 5e-5 for T5 models and 1e-5 for BARTs. We fixed the batch size at 32 when fine-tuning T5-BASE and BARTs. As for the extremely large T5-3B,
we configured a batch size of 64 to speed up convergence and utilised DeepSpeed to save memory. Linear learning rate decay was used for all models.

\paragraph{Machine Translation.} We evaluate the NLLB Translation Model at two distilled scales: 615M and 1.3B \citep{nllb}. Evaluation was done on an Nvidia GeForce RTX 2080 Ti and takes less than 10 minutes. The NLLB model is designed for many-to-many translation with low-resource language communities and is trained on a large corpus mined from the internet, rather than exclusively human aligned translations. We choose this model to give us an estimate of the performance of large scale translation products available to users.

\begin{figure*}
    \centering
    \includegraphics[width=\textwidth]{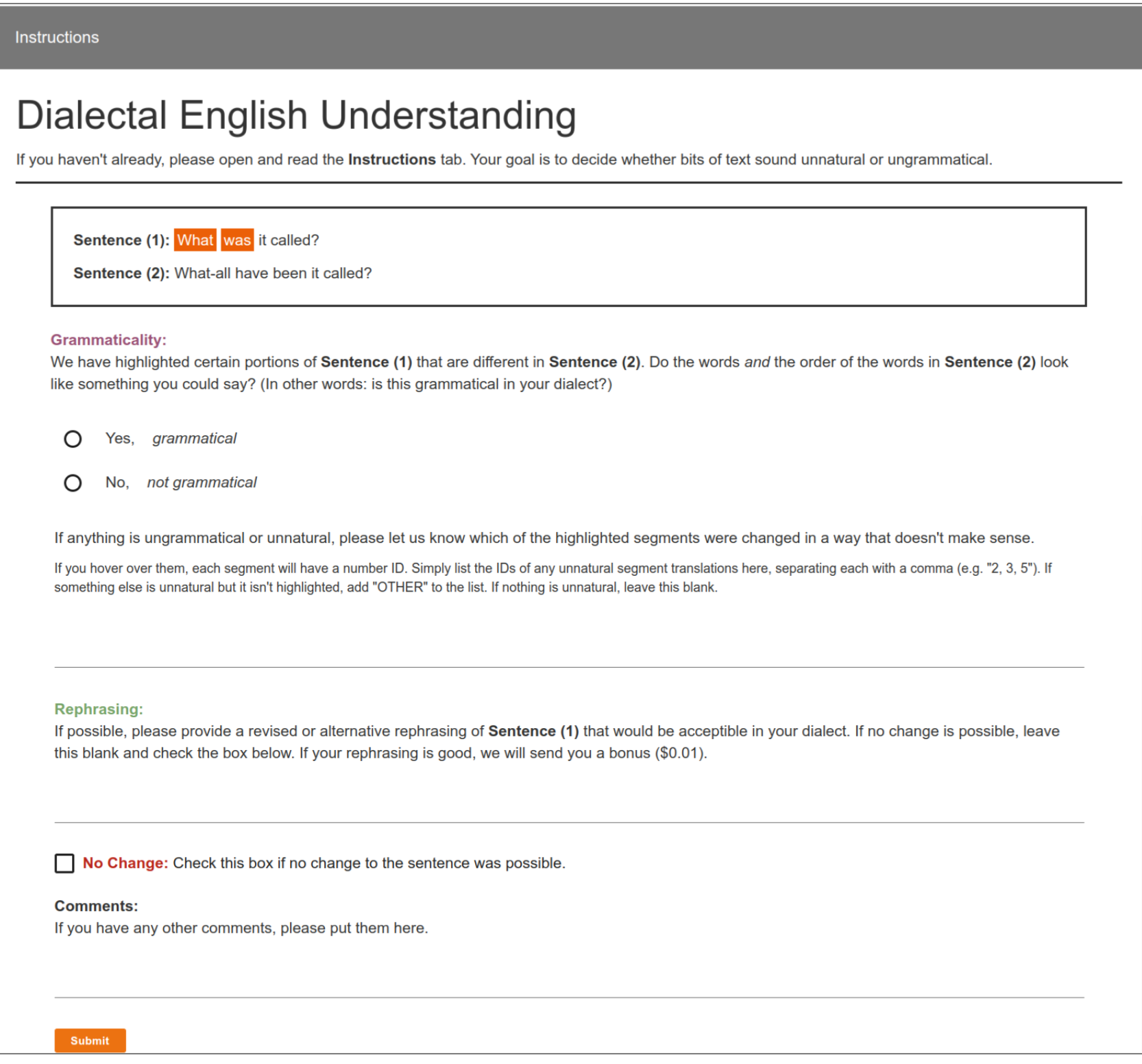}
    \caption{\textbf{MTurk Validation Task Interface.} Workers consider sentence pairs and evaluate whether the synthetic sentence is an acceptable dialectal form of the gloss given by the natural SAE sentence.}
    \label{fig:mturk_task}
\end{figure*}

\begin{table*}
\centering
\resizebox{0.8\textwidth}{!}{%
\def\arraystretch{1.15}
\begin{tabular}{lrrrrl}
\toprule
\textsc{Abbr} & \textsc{\# Feat.} & \textsc{\% Feat.} & \textsc{\# Val.} & \textsc{\% Val.} & \textsc{Dialect} \\ \midrule
AborE&89&83.2\%&57&53.3\%&Aboriginal English\\\hline
AppE&65&85.5\%&51&67.1\%&Appalachian English\\\hline
AusE&54&90.0\%&40&66.7\%&Australian English\\\hline
AusVE&47&83.9\%&34&60.7\%&Australian Vernacular English\\\hline
BahE&107&83.6\%&70&54.7\%&Bahamian English\\\hline
BlSAfE&95&88.0\%&71&65.7\%&Black South African English\\\hline
CamE&76&87.4\%&62&71.3\%&Cameroon English\\\hline
CFE&49&90.7\%&39&72.2\%&Cape Flats English\\\hline
ChIsE&47&94.0\%&33&66.0\%&Channel Islands English\\\hline
\gold{\textbf{ChcE}}&30&93.8\%&28&87.5\%&Chicano English\\\hline
CollAmE&57&83.8\%&44&64.7\%&Colloquial American English\\\hline
CollSgE&67&89.3\%&52&69.3\%&Colloquial Singapore English (Singlish)\\\hline
EAAVE&96&89.7\%&61&57.0\%&Earlier African American Vernacular English\\\hline
EA&46&85.2\%&32&59.3\%&East Anglian English\\\hline
FlkE&44&89.8\%&30&61.2\%&Falkland Islands English\\\hline
FijiE&39&88.6\%&36&81.8\%&Acrolectal Fiji English\\\hline
CollFijiE&95&85.6\%&68&61.3\%&Pure Fiji English (basilectal FijiE)\\\hline
GhE&58&92.1\%&49&77.8\%&Ghanaian English\\\hline
HKE&74&91.4\%&61&75.3\%&Hong Kong English\\\hline
\gold{\textbf{IndE}}&90&90.0\%&82&82.0\%&Indian English\\\hline
InSAfE&75&83.3\%&58&64.4\%&Indian South African English\\\hline
IrE&75&81.5\%&54&58.7\%&Irish English\\\hline
JamE&69&88.5\%&47&60.3\%&Jamaican English\\\hline
KenE&50&90.9\%&45&81.8\%&Kenyan English\\\hline
LibSE&86&84.3\%&58&56.9\%&Liberian Settler English\\\hline
MalE&68&89.5\%&57&75.0\%&Malaysian English\\\hline
MaltE&72&86.7\%&59&71.1\%&Maltese English\\\hline
ManxE&55&83.3\%&40&60.6\%&Manx English\\\hline
NZE&44&88.0\%&37&74.0\%&New Zealand English\\\hline
NfldE&84&85.7\%&53&54.1\%&Newfoundland English\\\hline
NigE&45&88.2\%&37&72.5\%&Nigerian English\\\hline
North&77&85.6\%&47&52.2\%&English dialects in the North of England\\\hline
O\&SE&30&81.1\%&19&51.4\%&Orkney and Shetland English\\\hline
OzE&56&86.2\%&43&66.2\%&Ozark English\\\hline
PakE&48&87.3\%&42&76.4\%&Pakistani English\\\hline
PhilE&92&85.2\%&71&65.7\%&Philippine English\\\hline
RAAVE&136&82.9\%&88&53.7\%&Rural African American Vernacular English\\\hline
ScE&44&80.0\%&30&54.5\%&Scottish English\\\hline
SEAmE&108&80.6\%&75&56.0\%&Southeast American enclave dialects\\\hline
SLkE&29&82.9\%&23&65.7\%&Sri Lankan English\\\hline
StHE&113&85.0\%&78&58.6\%&St. Helena English\\\hline
SE&46&93.9\%&33&67.3\%&English dialects in the Southeast of England\\\hline
SW&73&89.0\%&46&56.1\%&English dialects in the Southwest of England\\\hline
TznE&41&93.2\%&35&79.5\%&Tanzanian English\\\hline
TdCE&92&82.9\%&64&57.7\%&Tristan da Cunha English\\\hline
UAAVE&118&83.7\%&79&56.0\%&Urban African American Vernacular English\\\hline
UgE&65&86.7\%&52&69.3\%&Ugandan English\\\hline
WelE&76&80.9\%&53&56.4\%&Welsh English\\\hline
WhSAfE&41&83.7\%&35&71.4\%&White South African English\\\hline
WhZimE&61&88.4\%&46&66.7\%&White Zimbabwean English\\
\bottomrule
\end{tabular}
}
\caption{\textbf{\mval{} Implemented Dialects.} We've implemented 50 English dialects as shown in this table. We list the number of implemented features (\textsc{\# Feat}), the proportion of that dialect's catalogued eWAVE features implemented (\textsc{\% Feat}), the number of validated features (\textsc{\# Val}), and the proportion of that dialect's catalogued eWAVE features validated (\textsc{\% Val}). All dialects are at or above 80\% implemented and above 51.4\% validated. Gold \gold{\textbf{ChcE}} and \gold{\textbf{IndE}} indicate that we also release a Gold CoQA dev set in Chicano and Indian English.}
\label{tab:dialect_implementations}
\end{table*}

\begin{table*}
\centering
\resizebox{\textwidth}{!}{%
\def\arraystretch{1.15}
\begin{tabular}{rlL{75mm}L{75mm}r}
\toprule
{} &                     \textsc{Function} &                                            \textsc{SAE} &                                                \textsc{Transform} & \textsc{Val Acc.} \\
 &                              &                                                    &                                                    \\
\midrule
\ewave{1}  &                   \texttt{she\_inanimate\_objects} &                                                         It's a good bike &                                                                               She's a good bike \\ \hline
\ewave{2}  &                    \texttt{he\_inanimate\_objects} &              The driver's license? She wasn't allowed to renew it right? &                                    The driver's license? She wasn't allowed to renew 'im right? \\ \hline
\ewave{3}  &                       \texttt{referential\_thing} &  Christmas dinner? I think it's better to wait until after she's had it. &                  Christmas dinner? I think it's better to wait until after she's had the thing. & 100.0\\ \hline
\ewave{4}  &                         \texttt{pleonastic\_that} &                                                            It's raining. &                                                                                  Thass raining. \\ \hline
\ewave{5}  &                         \texttt{em\_subj\_pronoun} &                                  This old woman, she started packing up. &                                                         This old woman, 'em started packing up. \\ \hline
\ewave{6}  &                          \texttt{em\_obj\_pronoun} &                                                We just turned it around. &                                                                      We just turned 'im around. \\ \hline
\ewave{7}  &                  \texttt{me\_coordinate\_subjects} &                                            Michelle and I will come too. &                                                                  Me and Michelle will come too. \\ \hline
\ewave{8}  &              \texttt{myself\_coordinate\_subjects} &                                       My husband and I were late. &                                                         My husband and myself were late. \\ \hline
\ewave{9}  &                      \texttt{benefactive\_dative} &                                              I have to get one of those! &                                                                  I have to get me one of those! & \gold{\textbf{ChcE}} 100.0\\ \hline
\ewave{10} &                   \texttt{no\_gender\_distinction} &         Susan is a nurse but she does not like to put drips on patients. &                                 Susan is a nurse but he does not like to put drips on patients. & \gold{\textbf{IndE }} 97.4 \\ \hline
\ewave{11} &                  \texttt{regularized\_reflexives} &                                                         He hurt himself. &                                                                                He hurt hisself. & 100.0 \\ \hline
\ewave{12} &  \texttt{regularized\_reflexives\_object\_pronouns} &                                                       I'll do it myself. &                                                                              I'll do it meself. \\ \hline
\ewave{13} &             \texttt{regularized\_reflexives\_aave} &                                              They look after themselves. &                                                                     They look after theyselves. \\ \hline
\ewave{14} &                           \texttt{reflex\_number} &                                              We cannot change ourselves. &                                                                       We cannot change ourself. & \gold{\textbf{IndE}} 100.0 \\ \hline
\ewave{15} &                         \texttt{absolute\_reflex} &                           and he and the bull were tuggin' and wrestlin' &                                             and himself and the bull were tuggin' and wrestlin' & \gold{\textbf{IndE}} 100.0 \\ \hline
\ewave{16} &                         \texttt{emphatic\_reflex} &                                           They brought it by themselves. &                                                              They brought it by their own self. & \gold{\textbf{ChcE}} 100.0\\ \hline
\ewave{18} &                                    \texttt{my\_i} &                                                                  my book &                                                                                          I book \\ \hline
\ewave{19} &                                  \texttt{our\_we} &                                                                 our farm &                                                                                         we farm \\ \hline
\ewave{20} &                                  \texttt{his\_he} &                                                                 his book &                                                                                         he book \\ \hline
\ewave{21} &                              \texttt{their\_they} &                                                               their book &                                                                                       they book \\ \hline
\ewave{22} &                                \texttt{your\_you} &                                                                your book &                                                                                        you book \\ \hline
\ewave{23} &                              \texttt{your\_yalls} &                                                    Where are your books? &                                                                        Where are y'all's books? \\ \hline
\ewave{24} &                                 \texttt{his\_him} &                                                                 his book &                                                                                        him book \\ \hline
\ewave{25} &                              \texttt{their\_them} &                                                               their book &                                                                                       them book \\ \hline
\ewave{26} &                                   \texttt{my\_me} &                                                                  my book &                                                                                         me book & 100.0\\ \hline
\ewave{27} &                                  \texttt{our\_us} &                                                                 our book &                                                                                         us book \\ \hline
\ewave{29} &                                   \texttt{me\_us} &                                                        Show me the town! &                                                                               Show us the town! & 100.0 \\ \hline
\ewave{30} &                \texttt{non\_coordinated\_subj\_obj} &                                             Do you want to come with us? &                                                                    Do you want to come with we? \\ \hline
\ewave{31} &                \texttt{non\_coordinated\_obj\_subj} &                                                   They can ride all day. &                                                                          Them can ride all day. \\ \hline
\ewave{33} &                   \texttt{nasal\_possessive\_pron} &                                          her, his, our; hers, ours, ours &                                                           hern, hisn, ourn; hersn, oursn, ourns & 100.0\\ \hline
\ewave{34} &                                    \texttt{yall} &                                                                      you &  y'all & \gold{\textbf{ChcE}} \gold{\textbf{IndE}} 100.0\\ \hline
\ewave{35} &                                  \texttt{you\_ye} &                                     Sure it's no good to you in England. &                                                             Sure it's no good to ye in England. \\ \hline
\ewave{39} &                    \texttt{plural\_interrogative} &                                                                Who came? &                                                                                   Who-all came? & 99.7\\ \hline
\ewave{40} &               \texttt{reduplicate\_interrogative} &                                                      Who's coming today? &                                                                         Who-who's coming today? & \gold{\textbf{IndE}} 99.8 \\ \hline
\ewave{41} &                            \texttt{anaphoric\_it} &                  Things have become more expensive than they used to be. &                                           Things have become more expensive than it used to be. & \gold{\textbf{IndE}} 100.0\\ \hline
\ewave{42} &                     \texttt{object\_pronoun\_drop} &                                                 I got it from the store. &                                                                           I got from the store. & \gold{\textbf{IndE}} 98.1\\ \hline
\ewave{43} &               \texttt{null\_referential\_pronouns} &             When I come back from my work I just travel back to my home. &                                      When I come back from my work just travel back to my home. & \gold{\textbf{ChcE}} \gold{\textbf{IndE}} 93.2\\ \hline
\ewave{45} &                                 \texttt{it\_dobj} &                        As I explained to her, this is not the right way. &                                            As I explained it to her, this is not the right way. & \gold{\textbf{IndE}} 100.0\\ \hline
\ewave{46} &                       \texttt{it\_is\_referential} &                                                    It is very nice food. &                                                                              Is very nice food. \\ \hline
\ewave{47} &                   \texttt{it\_is\_non\_referential} &                                               Okay, it's time for lunch. &                                                                        Okay, is time for lunch. & \gold{\textbf{IndE}} 100.0\\
\bottomrule
\end{tabular}
}
\caption{\nameref{subsec:pronouns} (Section~\ref{subsec:pronouns})}
\label{tab:pronouns}
\end{table*}

\begin{table*}
\centering
\resizebox{\textwidth}{!}{%
\def\arraystretch{1.15}
\begin{tabular}{rlL{75mm}L{75mm}r}
\toprule
{} &                     \textsc{Function} &                                            \textsc{SAE} &                                                \textsc{Transform} & \textsc{Val Acc.} \\
 &                              &                                                    &                                                    \\
\midrule
\ewave{49} &                     \texttt{regularized\_plurals} &                                             wives, knives, lives, leaves &                                                       wifes, knifes, lifes, leafs & \gold{\textbf{IndE}} 99.6\\ \hline
\ewave{50} &                         \texttt{plural\_preposed} &                                                           shooting birds &                                                                shooting alla bird \\ \hline
\ewave{51} &                        \texttt{plural\_postposed} &                                                                 The boys &                                                                        Da boy dem \\ \hline
\ewave{55} &                       \texttt{mass\_noun\_plurals} &  furniture, machinery, equipment, evidence, luggage, advice, mail, staff &  furnitures, machineries, equipments, evidences, luggages, advices, mails, staffs & \gold{\textbf{IndE}} 100.0 \\ \hline
\ewave{56} &            \texttt{zero\_plural\_after\_quantifier} &                                               It's only five miles away. &                                                         It's only five mile away. & \gold{\textbf{ChcE}} \gold{\textbf{IndE}} 97.3 \\ \hline
\ewave{57} &                \texttt{plural\_to\_singular\_human} &                          The three girls there don't want to talk to us. &                                    The three girl there don't want to talk to us.& \gold{\textbf{IndE}} 100.0 \\ \hline
\ewave{58} &                             \texttt{zero\_plural} &                                              Some apartments are bigger. &                                                        Some apartment are bigger.& \gold{\textbf{IndE}} 100.0 \\ \hline
\ewave{59} &                      \texttt{double\_determiners} &                             This common problem of ours is very serious. &                                          This our common problem is very serious.& \gold{\textbf{IndE}} 100.0 \\ \hline
\ewave{60} &        \texttt{definite\_for\_indefinite\_articles} &                                                   She's got a toothache. &                                                           She's got the toothache & \gold{\textbf{IndE}} 99.6\\ \hline
\ewave{61} &        \texttt{indefinite\_for\_definite\_articles} &                                     The moon was very bright last night. &                                                A moon was very bright last night. & \gold{\textbf{IndE}} 100.0\\ \hline
\ewave{62} &                     \texttt{remove\_det\_definite} &                                                      He's in the office. &                                                                   He's in office. & \gold{\textbf{IndE}} 100.0\\ \hline
\ewave{63} &                   \texttt{remove\_det\_indefinite} &                                                Can I get a better grade? &                                                           Can I get better grade? & \gold{\textbf{IndE}} 99.0\\ \hline
\ewave{64} &                       \texttt{definite\_abstract} &                                             I stayed on until Christmas. &                                                  I stayed on until the Christmas. & \gold{\textbf{IndE}} 100.0 \\ \hline
\ewave{65} &                     \texttt{indefinite\_for\_zero} &                                           We received good news at last. &                                                  We received a good news at last. \\ \hline
\ewave{66} &                               \texttt{indef\_one} &                                         What happened? Oh, a dog bit me. &                                                What happened? Oh, one dog bit me. & \gold{\textbf{IndE}} 100.0 \\ \hline
\ewave{67} &     \texttt{demonstrative\_for\_definite\_articles} &                  They have two children. The elder girl is 19 years old. &                          They have two children. That elder girl is 19 years old. & \gold{\textbf{IndE}} 99.1\\ \hline
\ewave{68} &                              \texttt{those\_them} &                                I don't have any of those qualifications. &                                          I don't have any of them qualifications. \\ \hline
\ewave{70} &          \texttt{proximal\_distal\_demonstratives} &         this book that is right here vs. those books that are over there &                                               this here book vs. them there books & \gold{\textbf{ChcE}} 92.9\\ \hline
\ewave{71} &                 \texttt{demonstrative\_no\_number} &                                     These books are useful for my study. &                                               This books are useful for my study. & \gold{\textbf{IndE}} 98.8\\ \hline
\ewave{73} &                 \texttt{existential\_possessives} &                                                            I have a son. &                                                                     Son is there. \\ \hline
\ewave{74} &                    \texttt{possessives\_for\_post} &                                               This is my mother's house. &                                                  This is the house for my mother. \\ \hline
\ewave{75} &                     \texttt{possessives\_for\_pre} &                                Long time ago he was my sister's husband. &                                           Long time he was for my sister husband. \\ \hline
\ewave{76} &                      \texttt{possessives\_belong} &                                                       the woman's friend &                                                               woman belong friend \\ \hline
\ewave{77} &                           \texttt{null\_genitive} &                                                         my cousin's bike &                                                                    my cousin bike & \gold{\textbf{IndE}} 100.0\\ \hline
\ewave{78} &  \texttt{double\_comparative, double\_superlative} &                                        That is so much easier to follow. &                                            That is so much more easier to follow. & \gold{\textbf{IndE}} 100.0\\ \hline
\ewave{79} &                   \texttt{synthetic\_superlative} &                                       He is the most regular guy I know. &                                                  He is the regularest guy I know. & \gold{\textbf{IndE}} 100.0\\ \hline
\ewave{80} &                    \texttt{analytic\_superlative} &                                             one of the prettiest sunsets &                                                    one of the most pretty sunsets & \gold{\textbf{IndE}} 100.0\\ \hline
\ewave{81} &                               \texttt{more\_much} &                            The situation is more serious than I thought. &                                     The situation is much serious than I thought. & \gold{\textbf{IndE}} 100.0\\ \hline
\ewave{82} &                       \texttt{comparative\_as\_to} &                                           She is bigger than her sister. &                                                      She is bigger as her sister. \\ \hline
\ewave{84} &                        \texttt{comparative\_than} &                                 They like football more than basketball. &                                               They like football than basketball. \\ \hline
\ewave{85} &                    \texttt{comparative\_more\_and} &                                      He has more clothes than all of us. &                                                He has more clothes and all of us. \\ \hline
\ewave{86} &                             \texttt{zero\_degree} &           He is one of the most radical students that you can ever find. &                         He is one of the radical students that you can ever find. & \gold{\textbf{IndE}} 100.0\\ \hline
\ewave{87} &                             \texttt{adj\_postfix} &                                     A big and fresh fish is my favorite. &                                              A fish big and fresh is my favorite. \\
\bottomrule
\end{tabular}
}
\caption{\nameref{subsec:nounphrases} (Section~\ref{subsec:nounphrases})}
\label{tab:nounphrases}
\end{table*}

\begin{table*}
\centering
\resizebox{\textwidth}{!}{%
\def\arraystretch{1.15}
\begin{tabular}{rlL{75mm}L{75mm}r}
\toprule
{} &                     \textsc{Function} &                                            \textsc{SAE} &                                                \textsc{Transform} & \textsc{Val Acc.} \\
 &                              &                                                    &                                                    \\
\midrule
\ewave{88}  &                      \texttt{progressives} &                                    I like her hair style right now. &                                            I am liking her hair style. & \gold{\textbf{IndE}} 99.4\\ \hline
\ewave{95}  &                    \texttt{standing\_stood} &                                      He was standing on the corner. &                                            He was stood on the corner. \\ \hline
\ewave{96}  &  \texttt{that\_resultative\_past\_participle} &                         There is a car that broke down on the road. &                                There is a car broken down on the road. & 95.5 \\ \hline
\ewave{97}  &             \texttt{medial\_object\_perfect} &                                            He has written a letter. &                                               He has a letter written. \\ \hline
\ewave{98}  &                     \texttt{after\_perfect} &                                         She has just sold the boat. &                                          She's after selling the boat. \\ \hline
\ewave{99}  &   \texttt{simple\_past\_for\_present\_perfect} &                               I've eaten the food. So can I go now? &                                       I ate the food. So can I go now? & \gold{\textbf{ChcE}} 94.7 \\ \hline
\ewave{100} &          \texttt{present\_perfect\_for\_past} &                                            We were there last year. &                                            We've been there last year. & \gold{\textbf{IndE}} 99.9\\ \hline
\ewave{101} &           \texttt{present\_for\_exp\_perfect} &                               I've known her since she was a child. &                                      I know her since she was a child. & \gold{\textbf{IndE}} 100.0\\ \hline
\ewave{102} &                        \texttt{be\_perfect} &                                       They haven't left school yet. &                                           They're not left school yet. \\ \hline
\ewave{103} &                   \texttt{do\_tense\_marker} &                                   I knew some things weren't right. &                                  I did know some things weren't right. \\ \hline
\ewave{104} &                   \texttt{completive\_done} &                                     Sharon has read the whole book. &                                       Sharon done read the whole book. \\ \hline
\ewave{105} &              \texttt{completive\_have\_done} &                                             He has talked about me. &                                           He has done talked about me. \\ \hline
\ewave{106} &                  \texttt{irrealis\_be\_done} &  If you love your enemies, they will eat you alive in this society. &  If you love your enemies, they be done eat you alive in this society. & 100.0\\ \hline
\ewave{107} &                      \texttt{perfect\_slam} &                                                                 I have already told you not to mess up &                                                                    I slam told you not to mess up. \\ \hline
\ewave{108} &              \texttt{present\_perfect\_ever} &                                              I have seen the movie. &                                                  I ever see the movie. \\ \hline
\ewave{109} &                   \texttt{perfect\_already} &                                               Have you eaten lunch? &                                                   Did you eat already? \\ \hline
\ewave{110} &                 \texttt{completive\_finish} &                                                       I have eaten. &                                                          I finish eat. \\ \hline
\ewave{111} &                         \texttt{past\_been} &                                                         I told you. &                                                       I been told you. \\ \hline
\ewave{112} &                      \texttt{bare\_perfect} &                       We had caught the fish when the big wave hit. &                           We had catch the fish when the big wave hit. \\ \hline
\ewave{114} &                    \texttt{future\_sub\_gon} &                                               He will come with us. &                                                  He gon' come with us. \\ \hline
\ewave{115} &                  \texttt{volition\_changes} &                                                     You want to go. &                                                           You waan go. \\ \hline
\ewave{116} &                       \texttt{come\_future} &                                       I am about to cook your meal. &                                         I am coming to cook your meal. \\ \hline
\ewave{117} &        \texttt{present\_for\_neutral\_future} &       Next week, I will be leaving the States and going to Liberia. &                   Next week, I leaving the States, I going to Liberia. & \gold{\textbf{IndE}} 100.0\\ \hline
\ewave{118} &                          \texttt{is\_am\_1s} &                                                 I am going to town. &                                                     I's going to town. \\ \hline
\ewave{119} &                        \texttt{will\_would} &                                           I will meet him tomorrow. &                                             I would meet him tomorrow.& \gold{\textbf{IndE}} 100.0 \\ \hline
\ewave{120} &                          \texttt{if\_would} &                                  If I were you I would go home now. &                                 If I would be you I would go home now. \\
\bottomrule
\end{tabular}
}
\caption{\nameref{subsec:tenseaspect} (Section~\ref{subsec:tenseaspect})}
\label{tab:tenseaspect}
\end{table*}

\begin{table*}
\centering
\resizebox{\textwidth}{!}{%
\def\arraystretch{1.15}
\begin{tabular}{rlL{75mm}L{75mm}r}
\toprule
{} &                     \textsc{Function} &                                            \textsc{SAE} &                                                \textsc{Transform} & \textsc{Val Acc.}\\
 &                              &                                                    &                                                    \\
\midrule
\ewave{121} &               \texttt{double\_modals} &            We could do that. &                 We might could do that. & \gold{\textbf{ChcE  }} 91.8\\ \hline
\ewave{123} &              \texttt{present\_modals} &  I wish I could get the job. &               I wish I can get the job. & \gold{\textbf{IndE}} 100.0\\ \hline
\ewave{126} &  \texttt{finna\_future, fixin\_future} &      They're about to leave. &  They're fixin to leave town. & \gold{\textbf{ChcE  }} 92.3\\
\bottomrule
\end{tabular}
}
\caption{\nameref{subsec:mood} (Section~\ref{subsec:mood})}
\label{tab:mood}
\end{table*}

\begin{table*}
\centering
\resizebox{\textwidth}{!}{%
\def\arraystretch{1.15}
\begin{tabular}{rlL{75mm}L{75mm}r}
\toprule
{} &                     \textsc{Function} &                                            \textsc{SAE} &                                                \textsc{Transform} & \textsc{Val Acc.}\\
 &                              &                                                    &                                                    \\
\midrule
\ewave{128} &    \texttt{regularized\_past\_tense} &                     He caught the ball. &                      He catched the ball. & \gold{\textbf{ChcE}} \gold{\textbf{IndE}} 92.7\\ \hline
\ewave{129} &           \texttt{bare\_past\_tense} &                They came and joined us. &                  They come and joined us. \\ \hline
\ewave{130} &  \texttt{past\_for\_past\_participle} &                            He had gone. &                              He had went. & \gold{\textbf{ChcE}} 92.9 \\ \hline
\ewave{131} &     \texttt{participle\_past\_tense} &                               I saw it. &                                I seen it. & \gold{\textbf{IndE}} 100.0 \\ \hline
\ewave{132} &           \texttt{bare\_past\_tense} &  Here are things you ordered yesterday. &      Here are things you order yesterday. & \gold{\textbf{ChcE}} \gold{\textbf{IndE}} 87.7\\ \hline
\ewave{133} &               \texttt{double\_past} &          They didn't make it this time. &            They didn't made it this time. & \gold{\textbf{IndE}} 99.5\\ \hline
\ewave{134} &                     \texttt{a\_ing} &                    Where are you going? &                     Where are you a-goin? & 100.0\\ \hline
\ewave{135} &              \texttt{a\_participle} &              You've killed your mother. &              You've a-killed your mother. \\ \hline
\ewave{143} &         \texttt{transitive\_suffix} &                   You can see the fish. &                     You can see 'im fish. \\ \hline
\ewave{145} &                \texttt{got\_gotten} &   I hope you've got your topic already. &  I hope you've gotten your topic already. & 100.0\\ \hline
\ewave{146} &         \texttt{verbal\_ing\_suffix} &                        I can drive now. &                        I can driving now. & \gold{\textbf{IndE}} 100.0 \\ \hline
\ewave{147} &      \texttt{conditional\_were\_was} &                           If I were you &                              If I was you \\ \hline
\ewave{148} &          \texttt{serial\_verb\_give} &                  I bought rice for you. &                      I buy rice give you. \\ \hline
\ewave{149} &            \texttt{serial\_verb\_go} &         Grandfather sends us to school. &            Grandfather send us go school. & 100.0\\ \hline
\ewave{150} &                 \texttt{here\_come} &                    Bring the book here. &                 Take the book bring come. \\ \hline
\ewave{153} &              \texttt{give\_passive} &            John was scolded by his boss &                 John give his boss scold. \\
\bottomrule
\end{tabular}
}
\caption{\nameref{subsec:verbmorphology} (Section~\ref{subsec:verbmorphology})}
\label{tab:verbmorphology}
\end{table*}

\begin{table*}
\centering
\resizebox{\textwidth}{!}{%
\def\arraystretch{1.15}
\begin{tabular}{rlL{75mm}L{75mm}r}
\toprule
{} &                     \textsc{Function} &                                            \textsc{SAE} &                                                \textsc{Transform} & \textsc{Val Acc.}\\
 &                              &                                                    &                                                    \\
\midrule
\ewave{154} &            \texttt{negative\_concord} &                           I don't want any help. &                   I don't want no help. & \gold{\textbf{ChcE}} \gold{\textbf{IndE}} 92.9\\ \hline
\ewave{155} &                     \texttt{aint\_be} &                                 That isn't fair. &                        That ain't fair. & \gold{\textbf{ChcE}} 81.8\\ \hline
\ewave{156} &                   \texttt{aint\_have} &                          I hadn't seen them yet. &                  I ain't seen them yet. \\ \hline
\ewave{157} &            \texttt{aint\_before\_main} &                    something I didn't know about &            something I ain't know about \\ \hline
\ewave{158} &                        \texttt{dont} &                He doesn't always tell the truth. &         He don't always tell the truth. \\ \hline
\ewave{159} &               \texttt{never\_negator} &                                  He didn't come. &                          He never came. & 100.0 \\ \hline
\ewave{160} &        \texttt{no\_preverbal\_negator} &                I don't want any job or anything. &          I no want any job or anything. \\ \hline
\ewave{161} &       \texttt{not\_preverbal\_negator} &        The baby didn't eat food and cried a lot. &  The baby not ate food and cried a lot. \\ \hline
\ewave{162} &            \texttt{nomo\_existential} &       There is not any food in the refrigerator. &       No more food in the refrigerator. \\ \hline
\ewave{163} &                \texttt{wasnt\_werent} &                  John was there, but Mike wasn't &        John was there, but Mike weren't \\ \hline
\ewave{164} &          \texttt{invariant\_tag\_amnt} &  I believe I am older than you. Is that correct? &           I am older than you, amn't I? \\ \hline
\ewave{165} &   \texttt{invariant\_tag\_non\_concord} &          I believe you are ill. Is that correct? &                  You are ill, isn't it? & \gold{\textbf{IndE}} 99.1\\ \hline
\ewave{166} &    \texttt{invariant\_tag\_can\_or\_not} &                                   Can I go home? &          I want to go home, can or not? \\ \hline
\ewave{167} &  \texttt{invariant\_tag\_fronted\_isnt} &                      I can go there now can't I? &              Isn't, I can go there now? \\
\bottomrule
\end{tabular}
}
\caption{\nameref{subsec:negation} (Section~\ref{subsec:negation})}
\label{tab:negation}
\end{table*}

\begin{table*}
\centering
\resizebox{\textwidth}{!}{%
\def\arraystretch{1.15}
\begin{tabular}{rlL{75mm}L{75mm}r}
\toprule
{} &                     \textsc{Function} &                                            \textsc{SAE} &                                                \textsc{Transform} & \textsc{Val Acc.} \\
 &                              &                                                    &                                                    \\
\midrule
\ewave{170} &                   \texttt{uninflect} &                      He speaks English. &                                                            He speak English. & \gold{\textbf{ChcE}} \gold{\textbf{IndE}} 94.9\\ \hline
\ewave{171} &  \texttt{generalized\_third\_person\_s} &           Every Sunday we go to church. &                                              Every Sunday we goes to church. \\ \hline
\ewave{172} &           \texttt{existential\_there} &  There are two men waiting in the hall. &                                         There's two men waiting in the hall. & \gold{\textbf{ChcE}} \gold{\textbf{IndE}} 90.0\\ \hline
\ewave{173} &              \texttt{existential\_it} &        There's some milk in the fridge. &                                                It's some milk in the fridge. & \gold{\textbf{ChcE}} 87.5 \\ \hline
\ewave{174} &     \texttt{drop\_aux\_be\_progressive} &       You are always thinking about it. &                                                You always thinking about it. & \gold{\textbf{IndE}} 100.0\\ \hline
\ewave{175} &           \texttt{drop\_aux\_be\_gonna} &       He is gonna go home and watch TV. &                                               He gonna go home and watch TV. & \gold{\textbf{ChcE}} \gold{\textbf{IndE}} 83.3 \\ \hline
\ewave{176} &           \texttt{drop\_copula\_be\_NP} &                   He is a good teacher. &                                                           He a good teacher. \\ \hline
\ewave{177} &           \texttt{drop\_copula\_be\_AP} &                           She is smart. &                                                                   She smart. \\ \hline
\ewave{178} &     \texttt{drop\_copula\_be\_locative} &                         She is at home. &                                                                 She at home. \\ \hline
\ewave{179} &               \texttt{drop\_aux\_have} &                  I have seen it before. &                                                            I seen it before. & \gold{\textbf{IndE}} 100.0\\ \hline
\ewave{180} &                    \texttt{were\_was} &     You were hungry but he was thirsty. &  You was hungry but he was thirsty. OR: You were hungry but he were thirsty. \\
\bottomrule
\end{tabular}
}
\caption{\nameref{subsec:agreement} (Section~\ref{subsec:agreement})}
\label{tab:agreement}
\end{table*}

\begin{table*}
\centering
\resizebox{\textwidth}{!}{%
\def\arraystretch{1.15}
\begin{tabular}{rlL{75mm}L{75mm}r}
\toprule
{} &                     \textsc{Function} &                                            \textsc{SAE} &                                                \textsc{Transform} & \textsc{Val Acc.}\\
 &                              &                                                    &                                                    \\
\midrule
\ewave{186} &                   \texttt{who\_which} &                                                                           He's the man who looks after the cows. &                                                                 He's the man which looks after the cows. \\ \hline
\ewave{187} &                      \texttt{who\_as} &                                                                                       The man who was just here. &                                                                                The man as was just here. \\ \hline
\ewave{188} &                      \texttt{who\_at} &                                                                            This is the man who painted my house. &                                                                     This is the man at painted my house. \\ \hline
\ewave{189} &           \texttt{relativizer\_where} &  My father was one of the founders of the Underground Railroad, which helped the slaves to run away to the North &  My father was one o de founders o' de Underground Railroad where help de slaves to run way to de North. \\ \hline
\ewave{190} &                    \texttt{who\_what} &                                                                            This is the man who painted my house. &                                                                   This is the man what painted my house. \\ \hline
\ewave{191} &        \texttt{relativizer\_doubling} &                                                            But these, these little fellahs who had stayed before &                                              But these, these little fellahs that which had stayed befo' & \gold{\textbf{IndE}} 100.0\\ \hline
\ewave{192} &  \texttt{analytic\_whose\_relativizer} &                                                                             This is the man whose wife has died. &                      This is the man that his wife has died. OR: This is the man what his wife has died. \\ \hline
\ewave{193} &                  \texttt{null\_relcl} &                                                                             The man who lives there is friendly. &                                                                         The man lives there is friendly. & \gold{\textbf{ChcE}} \gold{\textbf{IndE}} 88.7\\ \hline
\ewave{194} &             \texttt{shadow\_pronouns} &                                                                     This is the house which I painted yesterday. &                                                          This is the house which I painted it yesterday. & \gold{\textbf{IndE}} 100.0 \\ \hline
\ewave{195} &             \texttt{one\_relativizer} &                                                              The cake that John buys is always very nice to eat. &                                                           The cake John buy one always very nice to eat. \\ \hline
\ewave{196} &   \texttt{correlative\_constructions} &                                                                               The ones I made are the good ones. &                                                                        The one I made, that one is good. \\ \hline
\ewave{197} &               \texttt{linking\_relcl} &                          Unless you are going to get 88, but some universities are not going to give those marks &                 Unless you are going to get 88 which some universities are not going to give those marks \\ \hline
\ewave{198} &        \texttt{preposition\_chopping} &                                                      You remember the swing that we all used to sit together on? &                                                 You remember the swing that we all used to sit together?& \gold{\textbf{IndE}} 100.0 \\ \hline
\ewave{199} &            \texttt{reduced\_relative} &                                                                       There is nothing like food cooked by Amma! &                                                                  There is nothing like Amma cooked food! \\
\bottomrule
\end{tabular}
}
\caption{\nameref{subsec:relativization} (Section~\ref{subsec:relativization})}
\label{tab:relativization}
\end{table*}

\begin{table*}
\centering
\resizebox{\textwidth}{!}{%
\def\arraystretch{1.15}
\begin{tabular}{rlL{75mm}L{75mm}r}
\toprule
{} &                     \textsc{Function} &                                            \textsc{SAE} &                                                \textsc{Transform} & \textsc{Val Acc.} \\
 &                              &                                                    &                                                    \\
\midrule
\ewave{200} &          \texttt{say\_complementizer} &                            We hear that you were gone to the city. &                                                                 We hear say you gone to the city. \\ \hline
\ewave{201} &          \texttt{for\_complementizer} &          You mean your mother allows you to bring over boyfriends? &                                        You mean your mother allows you for bring over boyfriends? \\ \hline
\ewave{202} &               \texttt{for\_to\_pupose} &  We always had gutters in the winter time to drain the water away. &                             We always had gutters in the winter time for to drain the water away. \\ \hline
\ewave{203} &                      \texttt{for\_to} &                        He had the privilege to turn on the lights. &  He had the privilege for to turn on the lights. OR: He had the privilege for turn on the lights. & 100.0\\ \hline
\ewave{204} &            \texttt{what\_comparative} &                                             I'm taller than he is. &                                                                       I'm taller than what he is. & \gold{\textbf{IndE}} 100.0\\ \hline
\ewave{205} &             \texttt{existential\_got} &                                    There's no water in the toilet. &                                                                       Got no water in the toilet. & 100.0\\ \hline
\ewave{206} &        \texttt{existential\_you\_have} &         There are some people who don't give a damn about animals. &                                        You have some people they don't give a damn about animals. & \gold{\textbf{IndE}} 100.0\\ \hline
\ewave{207} &  \texttt{that\_infinitival\_subclause} &                                       He wanted me to go with him. &                                                              He wanted that I should go with him. & \gold{\textbf{IndE}} 100.0\\ \hline
\ewave{208} &                 \texttt{drop\_inf\_to} &                                     They were allowed to call her. &                                                                       They were allowed call her. & 100.0\\ \hline
\ewave{209} &               \texttt{to\_infinitive} &                                                  He made me do it. &                                                                              He made me to do it. & \gold{\textbf{IndE}} 100.0\\ \hline
\ewave{210} &                  \texttt{bare\_ccomp} &              When mistress started whooping her, she sat her down. &                                                When mistress started whoop her, she sat her down. \\
\bottomrule
\end{tabular}
}
\caption{\nameref{subsec:complementation} (Section~\ref{subsec:complementation})}
\label{tab:complementation}
\end{table*}

\begin{table*}
\centering
\resizebox{\textwidth}{!}{%
\def\arraystretch{1.15}
\begin{tabular}{rlL{75mm}L{75mm}r}
\toprule
{} &                     \textsc{Function} &                                            \textsc{SAE} &                                                \textsc{Transform} & \textsc{Val Acc.}\\
 &                              &                                                    &                                                    \\
\midrule
\ewave{211} &      \texttt{clause\_final\_though\_but} &                          There's nothing wrong with this box though. &                               There's nothing wrong with this box, but. \\ \hline
\ewave{212} &      \texttt{clause\_final\_really\_but} &                           I don't know what else she can do, really. &                                 I don't know what else she can do, but. \\ \hline
\ewave{213} &          \texttt{chaining\_main\_verbs} &                        If you stay longer, they have to charge more. &                                  Stay longer, they have to over-charge. \\ \hline
\ewave{214} &    \texttt{corr\_conjunction\_doubling} &  Despite being instructed on what to do, he still made some misakes. &  Despite being instructed on what to do still yet he made some misakes.& \gold{\textbf{IndE}} 100.0 \\ \hline
\ewave{215} &  \texttt{subord\_conjunction\_doubling} &                      Although you are smart, you are not appreciated &                     Although you are smart, but you are not appreciated \\
\bottomrule
\end{tabular}
}
\caption{\nameref{subsec:adverbialsubordination} (Section~\ref{subsec:adverbialsubordination})}
\label{tab:adverbialsubordination}
\end{table*}

\begin{table*}
\centering
\resizebox{\textwidth}{!}{%
\def\arraystretch{1.15}
\begin{tabular}{rlL{75mm}L{75mm}r}
\toprule
{} &                     \textsc{Function} &                                            \textsc{SAE} &                                                \textsc{Transform} & \textsc{Val Acc.}\\
 &                              &                                                    &                                                    \\
\midrule
\ewave{216} &   \texttt{null\_prepositions} &                             I'm going to town. &                               I'm going town. & \gold{\textbf{IndE}} 99.7\\ \hline
\ewave{220} &  \texttt{degree\_adj\_for\_adv} &                    That's really nice and cold &                     That's real nice and cold &\gold{\textbf{ChcE}} \gold{\textbf{IndE}} 99.4\\ \hline
\ewave{221} &    \texttt{flat\_adj\_for\_adv} &                          She speaks so softly. &                           She speaks so soft. & \gold{\textbf{ChcE}} \gold{\textbf{IndE}} 86.7 \\ \hline
\ewave{222} &             \texttt{too\_sub} &  They are very nice. We had a good time there. &  They are too nice. We had a good time there. \\
\bottomrule
\end{tabular}
}
\caption{\nameref{subsec:adverbsprepositions} (Section~\ref{subsec:adverbsprepositions})}
\label{tab:adverbsprepositions}
\end{table*}

\begin{table*}
\centering
\resizebox{\textwidth}{!}{%
\def\arraystretch{1.15}
\begin{tabular}{rlL{75mm}L{75mm}r}
\toprule
{} &                     \textsc{Function} &                                            \textsc{SAE} &                                                \textsc{Transform} & \textsc{Val Acc.} \\
 &                              &                                                    &                                                    \\
\midrule
\ewave{223} &                        \texttt{clefting} &  A lot of them are looking for more land. &  It's looking for more land a lot of them are. & 100.0\\ \hline
\ewave{224} &                   \texttt{fronting\_pobj} &           I drive to town every Saturday. &                To town every Saturday I drive. & \gold{\textbf{IndE}} 99.5\\ \hline
\ewave{226} &              \texttt{negative\_inversion} &                         Nobody showed up. &                         Didn't nobody show up. & 100.0\\ \hline
\ewave{227} &      \texttt{inverted\_indirect\_question} &   I'm wondering what you are going to do. &        I'm wondering what are you going to do. & \gold{\textbf{ChcE}} \gold{\textbf{IndE}} 91.2\\ \hline
\ewave{228} &                     \texttt{drop\_aux\_wh} &                       When is she coming? &                               When she coming? & \gold{\textbf{IndE}} 99.8\\ \hline
\ewave{229} &                     \texttt{drop\_aux\_yn} &                     Do you get the point? &                             You get the point? & \gold{\textbf{IndE}} 99.9\\ \hline
\ewave{230} &              \texttt{doubly\_filled\_comp} &                             Who ate what? &                            What who has eaten? \\ \hline
\ewave{231} &  \texttt{superlative\_before\_matrix\_head} &          The thing I like most is apples. &               The most thing I like is apples. \\ \hline
\ewave{232} &                \texttt{double\_obj\_order} &                 She would teach it to us. &                             She'd teach us it. & \gold{\textbf{IndE}} 100.0\\ \hline
\ewave{234} &             \texttt{acomp\_focusing\_like} &                      It was really cheap. &                      It was like really cheap. & \gold{\textbf{ChcE}} \gold{\textbf{IndE}} 91.2\\ \hline
\ewave{235} &                  \texttt{quotative\_like} &              And my friend said "No way!" &               And my friend was like "No way!" \\
\bottomrule
\end{tabular}
}
\caption{\nameref{subsec:discoursewordorder} (Section~\ref{subsec:discoursewordorder})}
\label{tab:discoursewordorder}
\end{table*}

\end{document}